\documentclass[review]{elsarticle}

\usepackage{lineno,hyperref}
\modulolinenumbers[5]

\journal{  }%Advances In Water Resources}

\usepackage[utf8]{inputenc} % allow utf-8 input
\usepackage[T1]{fontenc}    % use 8-bit T1 fonts
\usepackage{hyperref}       % hyperlinks
\usepackage{url}                 % simple URL typesetting
\usepackage{booktabs}       % professional-quality tables
\usepackage{amsfonts}       % blackboard math symbols
\usepackage{nicefrac}       % compact symbols for 1/2, etc.
\usepackage{microtype}      % microtypography
\usepackage{lipsum}
\usepackage{color,soul}  % for comment hl
\usepackage{amsmath}  % for align % Used for the boldsymbol.
\usepackage{graphicx} % use graphics

\usepackage{subfig} % use multiple figures

\usepackage{multirow}

\renewcommand{\vec}[1]{\boldsymbol{#1}}
\begin{document}

\begin{frontmatter}

\title{Physics-Informed Neural Networks for Multiphysics Data Assimilation with Application to Subsurface Transport}


\author[PNNL]{QiZhi He}

\author[PNNL]{David Brajas-Solano}

\author[INTERA]{Guzel Tartakovsky}

\author[PNNL]{Alexandre M. Tartakovsky\corref{mycorrespondingauthor}}
\cortext[mycorrespondingauthor]{Corresponding author}
\ead{Alexandre.Tartakovsky@pnnl.gov}

\address[PNNL]{Pacific Northwest National Laboratory Richland, WA 99354}
\address[INTERA]{INTERA Incorporated, Richland, WA 99354}

\begin{abstract}
Data assimilation for parameter and state estimation in subsurface transport problems remains a significant challenge due to the sparsity of measurements, the heterogeneity of porous media, and the high computational cost of forward numerical models. We present a physics-informed deep neural networks (DNNs) machine learning method for estimating space-dependent hydraulic conductivity, hydraulic head, and concentration fields from sparse measurements. In this approach, we employ individual DNNs to approximate the unknown parameters (e.g., hydraulic conductivity) and states (e.g., hydraulic head and concentration) of a physical system, and  jointly train these DNNs by minimizing the loss function that consists of the governing equations residuals in addition to the error with respect to measurement data. We apply this approach to assimilate conductivity, hydraulic head, and concentration measurements for joint inversion of the conductivity, hydraulic head, and concentration fields in a steady-state advection--dispersion problem. We study the accuracy of the physics-informed DNN approach with respect to data size, number of variables (conductivity and head versus conductivity, head, and concentration), DNNs size, and DNN initialization during training. We demonstrate that the physics-informed DNNs are significantly more accurate than standard data-driven DNNs when the training  set consists of sparse data. We also show that the accuracy of parameter estimation increases as additional variables are inverted jointly.
\end{abstract}

\begin{keyword}
Physics-informed Deep Neural Networks\sep data assimilation \sep parameter estimation \sep inverse problems
\ tracer test 
\end{keyword}

\end{frontmatter}

%\linenumbers

\section{Introduction}\label{sec:intro}

Modeling of transport in heterogeneous porous media is a part of many environmental and engineering applications, including hydrocarbon recovery~\cite{Hartmann1999}, hydraulic fracking~\cite{hubbert1972mechanics}, exploitation of geothermal energy~\cite{Barbier2002}, geologic disposal of radioactive waste~\cite{Helton1993}, and groundwater contamination assessment~\cite{Rajib2017a}. Numerical models of transport in porous media require deterministic or statistical knowledge of the subsurface properties (e.g., hydraulic conductivity) and initial and boundary conditions~\cite{bear2010modeling,White1995}. However, because of heterogeneity and data sparsity, the parameters of natural systems are often unknown. Despite significant research in inverse methods ~\cite{Hoeksema1984,Vrugt2008a,Rajib2017a,Rajabi2018}, parameter estimation at a resolution required for accurate modeling of transport processes remains a challenge.

Parameter estimation is complicated by the fact that the parameters of interest (e.g., hydraulic conductivity) are difficult to measure directly. Most inverse (parameter estimation) methods  use indirect measurements (in addition to direct measurements) to estimate parameters. Data assimilation (or model--data integration) has been well recognized as an effective technique to reduce predictive uncertainties and improve model accuracy. Data assimilation is a process where model parameters and system states are updated using measurements and governing equations~\cite{Rajib2017a,Evensen1994,Rayner2016}. Data assimilation has been used in many fields, including atmospheric and oceanic sciences~\cite{Evensen2003,houtekamer2001sequential}, hydrology ~\cite{christakos2005methodological,Vrugt2008a}, subsurface transport  \cite{Rajib2017a,Rajabi2018,Zheng2019}, and uncertainty quantification~\cite{vrugt2005improved,Liu2007}. What makes subsurface flow and transport applications challenging is that the inverse problems for the subsurface flow and transport equations are highly nonlinear and the states and parameters are non-Gaussian~\cite{Zheng2019}. The non-linearity presents a challenge for both the direct inverse methods and  Bayesian parameter estimation methods.

Recent advances in machine learning (ML) methods and automatic differentiation (including software infrastructures such as TensorFlow \cite{ramsundar2018tensorflow}) has made them potentially powerful tools for parameter estimation and data assimilation. For example, Schmidt and Lipson~\cite{Bongard2007} applied symbolic regression to learn conservation laws, and  Brunton et al.~\cite{brunton2016discovering} used sparse regression to discover equations of nonlinear dynamics directly from data. Physics-informed deep neural networks (PINNs) were used to learn solutions and parameters in partial and ordinary differential equations \cite{Raissi2019,Lagaris1997,Weinan2018,Raissi2019}. Recently, PINNs were extended for inverse problems associated with partial differential equations with random coefficients (e.g., to estimate the space-dependent hydraulic conductivity using sparse measurements of conductivity and hydraulic head) \cite{Tartakovsky2018}. Deep neural networks (DNNs) were also combined with the finite element method (FEM) for estimating parameters in partial differential equations (PDEs) given that the PDE states are fully known 
\cite{xu2019neural}.

In this study, we extend the PINN-based parameter estimation method of~\cite{Tartakovsky2018} to assimilate measurements of hydraulic conductivity, hydraulic head, and solute concentration. In the PINN method, we use the Darcy and advection--dispersion equations jointly with data to train DNNs that represent space-dependent conductivity, head, and concentration fields.
During training of the DNNs, the governing equations and the associated boundary conditions are enforced at the collocation points over the domain. We demonstrate that for sparse data, PINNs significantly improve the accuracy of parameter and state estimation as compared to standard DNNs trained with data only. 

This paper is organized as follows. In Section \ref{sec:prelim}, we briefly review the feed-forward fully-connected deep networks and  the  PINN approach. The extension of PINN to the assimilation of multiphysics data is presented in Section \ref{sec:pinn-mul}. Performance of PINNs, including the dependence of the estimation errors on the number of measurements, is given in Section \ref{sec:Result}. The effect of the neural network size and the conductivity field complexity on the accuracy of the parameter estimation is discussed in Section \ref{sec:diss}. Conclusions are given in Section \ref{sec:conclusions}.

\section{Physics-informed neural networks}\label{sec:prelim}
In this section, we first give a brief overview of feed-forward fully-connected DNNs as a function approximation and regression tool~\cite{Hornik1989}.  Next, we describe the PINN framework that enforces physics in the regression and enables assimilation of different types of data. 

\subsection{DNN approximation}\label{sec:DNN-appr}
\begin{figure}[ht!]
	\centering
	\includegraphics[angle=0,width=4in]{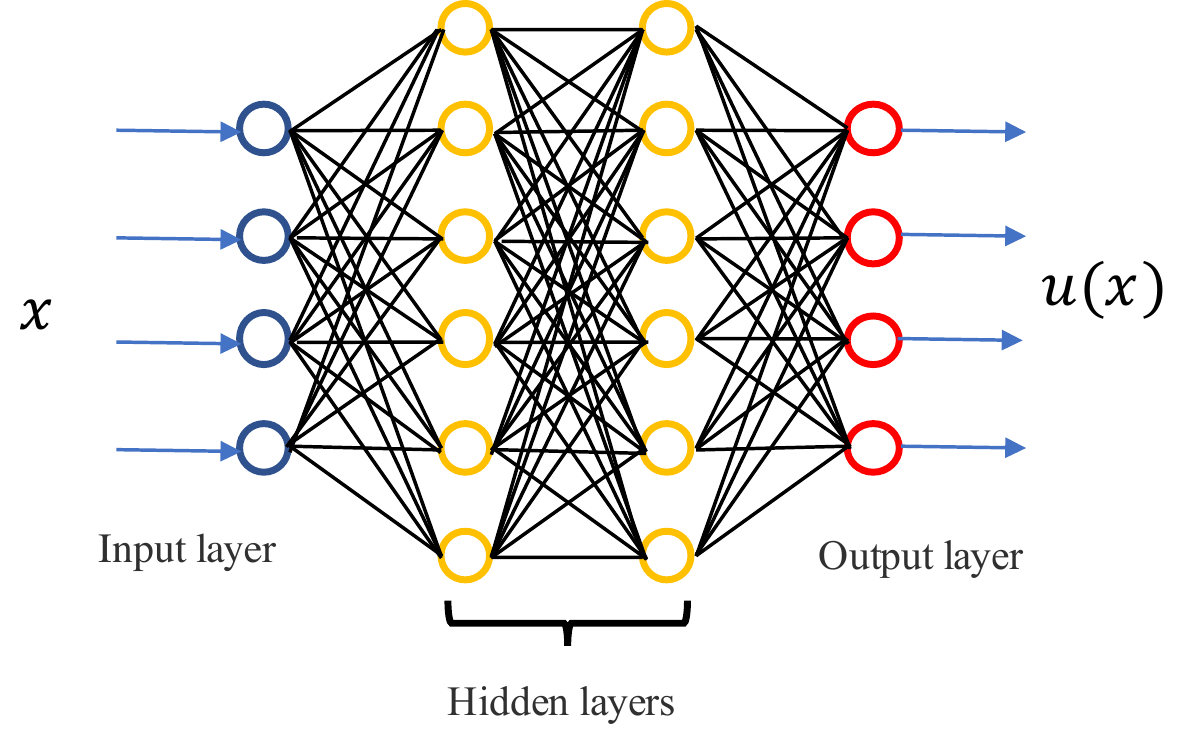}
	\caption{Feed-forward DNN.}
	\label{fig:DNN}
\end{figure}

We employ DNNs to approximate and regress a function $\vec{u}(\vec{x})$ using sparse measurements of $\vec{u}$.
We use a  fully-connected feed-forward network architecture known as multilayer perceptrons, where the basic computing units (neurons) are stacked in layers, as shown in Figure \ref{fig:DNN}. The first layer is called the input layer, and the last layer is the output layer, while all the intermediate layers are known as hidden layers. The DNN approximation of $\vec{u}(\vec{x})$ is given as:
\begin{equation}
\vec{u}(\vec{x}) \approx \hat{\vec{u}}(\vec{x};\theta)  = \vec{y}_{n_l+1} (\vec{y}_{n_l}(...(\vec{y}_2(\vec{x}))),
\end{equation}
where $\hat{(\cdot)}$ denotes a DNN approximation, and  
\begin{equation}
\begin{split}
\vec{y}_2(\vec{x}) &= \sigma(\vec{W}_1 \vec{x} + \vec{b}_1) \\
\vec{y}_3(\vec{y}_2) &= \sigma(\vec{W}_2 \vec{y}_2 + \vec{b}_2)\\
... \\
\vec{y}_{n_l} (\vec{y}_{n_l-1}) &= \sigma(\vec{W}_{n_l-1} \vec{y}_{n_l-1} + \vec{b}_{n_l-1})\\
\vec{y}_{n_l+1} (\vec{y}_{n_l}) &= \sigma(\vec{W}_{n_l} \vec{y}_{n_l} + \vec{b}_{n_l}).
\end{split}
\end{equation}
Here, $n_l$ denotes the number of hidden layers, $\sigma$ is the activation function, $\vec{x} \in \mathbb{R}^d$ denotes the input ($d$ is the number of spatial dimensions), $\vec{y}_{n_l+1} $ is the output vector, and $\theta$ is the vector of the weights and biases parameters:
\begin{equation}
\theta = \{\vec{W}_1, \vec{W}_2, ..., \vec{W}_{n_l}, \vec{b}_1, \vec{b}_2, ..., \vec{b}_{n_l}\}.
\end{equation}
In the ``data-driven'' approach, $\theta$ is estimated by minimizing the loss function  

\begin{equation}\label{loss_data}
\theta = \text{arg}\min \limits_\theta  \sum_{\vec{x} \in \mathcal{T}_u} (\hat{\vec{u}} (\vec{x}; \theta) - \vec{u}^*(\vec{x}))^2,
\end{equation}
where  $\mathcal{T}_u = \{\vec{x}_1,\vec{x}_2,...,\vec{x}_{|\mathcal{T}_u|} \}\subset \Omega$ denotes a set of measurement locations, $\Omega$ is the domain of function $\vec{u}$, and $\vec{u}^*(\vec{x}), \vec{x} \in \mathcal{T}_u $ are the measured values of $\vec{u}$ at these locations.

Some of the commonly used activation functions include logistic sigmoid, hyperbolic tangent, ReLu, and leaky ReLu. As the objective of this study is to approximate differentiable functions (space-dependent parameters and state variables in PDEs), we adopt the hyperbolic tangent activation function $\sigma(x) = \tanh(x)$, which is infinitely differentiable.

\subsection{Enforcing physics constraints}\label{sec:pinn}

 Given a sufficiently large number of hidden layers, DNNs have excellent representative properties but require a lot of data to train them. This creates a challenge in applying DNNs to subsurface problems where measurements are usually sparse. For the purpose of this work, we define sparse measurements as measurements that do not sufficiently cover the computational domain to accurately determine $\theta$ from the unconstrained minimization problem (\ref{loss_data}). In \cite{Tartakovsky2018}, we demonstrated that the Darcy law can be used as a constraint for solving the minimization problem (\ref{loss_data}) and learning hydraulic conductivity from sparse measurements of conductivity and hydraulic head using the PINN approach~\cite{Raissi2019}.

In the rest of this section, we extend the PINN parameter estimation method of \cite{Tartakovsky2018} to a data assimilation problem where different types of measurements (e.g., conductivity, hydraulic head, and concentration measurements) are used to estimate parameters (i.e., conductivity) and states (i.e., head and concentration fields).  Consider a system of PDEs forming the boundary value problem defined on the domain $\Omega \subset \mathbb{R}^d $ with the boundary $\partial \Omega$:
\begin{equation}\label{eq:model}
	\begin{split}
		\mathcal{L} (\vec{u}(\vec{x});\vec{p}(\vec{x})) &= 0, \quad \vec{x} \in \Omega\\
		\mathcal{B} (\vec{u}(\vec{x});\vec{p}(\vec{x})) & = 0, \quad \vec{x} \in \partial \Omega
	\end{split}
\end{equation}
where $\vec{u}$ is the (unknown) solution vector, $\vec{p}$ is the (unknown) system parameter vector, $\mathcal{L}$ denotes the known (nonlinear) differential operator, and the operator $\mathcal{B}$ expresses arbitrary boundary conditions associated with the problem. The boundary conditions can be of the Dirichlet and Neumann types applied on $\partial_D \Omega$ and $\partial_N \Omega$, respectively, such that $\partial_D \Omega \cup \partial_N \Omega = \partial \Omega$ and $\partial_D \Omega \cap \partial_N \Omega = \emptyset$.

Following \cite{Tartakovsky2018}, the DNNs are used to approximate unknown states and parameters,  $\vec{u}(\vec{x}) \approx \hat{\vec{u}} (\vec{x};\theta)$ and $\vec{p}(\vec{x}) \approx \hat{\vec{p}} (\vec{x};\gamma)$, $\vec{x} \in \Omega$.
% , with the "trainable" parameters $\theta$ and $\gamma$ as analogous to the generalized coefficients in standard numerical methods for PDEs. 
 To determine these parameters, we minimize the loss function $J(\theta,\gamma)$ with physics-informed penalty terms:
 \begin{equation}\label{minimization}
(\theta,\gamma ) = \text{arg}\min \limits_{\theta,\gamma} J(\theta,\gamma). 
\end{equation}
where
\begin{equation}\label{eq:loss_pinn}
\begin{split}
J(\theta,\gamma) = J_d (\theta,\gamma) +  \omega_f J_f(\theta,\gamma) + \omega_b J_b(\theta,\gamma).
\end{split}
\end{equation}
 Here, $J_d (\theta,\gamma)$ is the loss due to a mismatch with the data (i.e., the measurements of $\vec{u}$ and $\vec{p}$):
\begin{equation}
J_d (\theta,\gamma)  = \frac{1}{|\mathcal{T}_{u}|} \sum_{\vec{x} \in \mathcal{T}_u} (\hat{\vec{u}} (\vec{x}; \theta) - \vec{u}^*(\vec{x}))^2
+ \frac{1}{|\mathcal{T}_{p}|} \sum_{\vec{x} \in \mathcal{T}_p} (\hat{\vec{p}} (\vec{x}; \theta) - \vec{p}^*(\vec{x}))^2,
\end{equation} 
$J_f(\theta,\gamma)$ is the loss due to mismatch with the governing PDEs $\mathcal{L} (\vec{u}(\vec{x});\vec{p}(\vec{x}))= 0$: 
\begin{equation}
J_f(\theta,\gamma)  = \frac{1}{|\mathcal{T}_f|} \sum_{\vec{x} \in \mathcal{T}_f} (\mathcal{L} (\hat{\vec{u}}(\vec{x}; \theta); \hat{\vec{p}}(\vec{x},\gamma)))^2, 
\end{equation}
and
 $J_b(\theta,\gamma)$ is the loss due to mismatch with the boundary conditions $\mathcal{B} (\vec{u}(\vec{x});\vec{p}(\vec{x})) = 0$:
\begin{equation}
J_b(\theta,\gamma)  = \frac{1}{|\mathcal{T}_b|} \sum_{\vec{x} \in \mathcal{T}_b} (\mathcal{B} (\hat{\vec{u}}(\vec{x}; \theta); \hat{\vec{p}}(\vec{x};\gamma)))^2.
\end{equation}
In \eqref{eq:loss_pinn}, $\omega_f$, and $\omega_b$ are weights that determine how strongly mismatch with the governing PDEs and boundary conditions are penalized relative to data mismatch. In this work, we assume that the measurements and physics model are exact and set $\omega_f = \omega_b =1$. 
The sets $\mathcal{T}_u = \{\vec{x}_1,\vec{x}_2,...,\vec{x}_{|\mathcal{T}_u|} \}\subset \Omega$ and $\mathcal{T}_p = \{\vec{x}_1,\vec{x}_2,...,\vec{x}_{|\mathcal{T}_p|} \}\subset \Omega$ denote  the measurement locations of $\vec{u}$ and $\vec{p}$, respectively,  and $\vec{u}^*(\vec{x}), \vec{x} \in \mathcal{T}_u $ and $\vec{p}^*(\vec{x}), \vec{x} \in \mathcal{T}_p$ are the measured values of $\vec{u}$ and $\vec{p}$ at these locations.
The sets $\mathcal{T}_f = \{\vec{x}_1,\vec{x}_2,...,\vec{x}_{|\mathcal{T}_f|} \} \subset \Omega$ and $\mathcal{T}_b = \{\vec{x}_1,\vec{x}_2,...,\vec{x}_{|\mathcal{T}_b|} \} \subset \partial \Omega$ denote locations of the ``residual'' points where $J_f (\theta,\gamma)$ and $J_b(\theta,\gamma)$ are, respectively, minimized. The penalty terms $J_f (\theta,\gamma)$ and $J_b(\theta,\gamma)$ force the DNN approximations of $\vec{u}$ and $\vec{p}$ to satisfy the governing equation \eqref{eq:model} at the residual points. Note that while it is preferable to enforce physics over the whole domain, the computational cost of estimating and minimizing the loss function (\ref{eq:loss_pinn}) increases with the number of residual points. In this work, we demonstrate convergence of the solution of (\ref{minimization}) with an increasing number of residual points, meaning that the DNNs $\hat{\vec{u}}(\vec{x}; \theta)$ and $\hat{\vec{p}}(\vec{x};\gamma)$ can be accurately trained using a finite number of  residual points.  Similar convergence results for solving partial differential equations with the PINN method were also observed in ~\cite{Raissi2019,Lu2019,Tartakovsky2018,Lagaris1997,Rudd2013}.

The loss  $J_f(\theta,\gamma)$ is evaluated by computing spatial derivatives of $\hat{\vec{u}}(\vec{x}; \theta)$ and $\hat{\vec{p}}(\vec{x};\gamma)$ using automatic differentiation (AD)~\cite{Baydin2015}. AD is also used to evaluate the normal derivative $\vec{n} \cdot \nabla$ in the Neumann boundary condition in the loss $J_b(\theta,\gamma)$ (see details in Section \ref{sec:pinn-mul}). AD is implemented in most ML libraries, including TensorFlow and Pytorch \cite{paszke2017automatic}, where it is mainly used to compute derivatives with respect to the DNN weights (i.e., $\theta$ and $\gamma$). In the PINN method, AD allows the implementation of any PDE and boundary condition constraints without numerically discretizing and solving the PDEs.

Another benefit of enforcing PDE constraints via the penalty term $J_f(\theta,\gamma)$ is that we can use the corresponding weight $\omega_f$ to account for the ``fidelity'' of the PDE model. For example, we can assign a smaller weight to a low-fidelity PDE model. In general, the number of unknown parameters in $\theta$ and $\gamma$ is much larger than the number of measurements, and training of the DNNs requires regularization. One can consider the losses $J_b(\theta,\gamma)$ and $J_f(\theta,\gamma)$ in the minimization problem (\ref{minimization}) as a physics-informed regularization terms \cite{Tartakovsky2018,Nabian2018}. 

%%%%%%%%%%%%%%%%%%%%%%%%%%%%%%%%%%%%%%%%%%%%%%%%%%%%
%%%     PINN for Multiphysics Problems     %%%
%%%%%%%%%%%%%%%%%%%%%%%%%%%%%%%%%%%%%%%%%%%%%%%%%%%%
\section{PINN for data assimilation and parameter estimation in subsurface transport problems}\label{sec:pinn-mul}

For sparsely  sampled systems, data assimilation can significantly improve the accuracy of parameter and state estimation. 
Here, we assume that the sparse steady-state measurements of a synthetic tracer test in a heterogeneous porous domain $\Omega = [0,L_1]\times[0,L_2]$ are available, where the solute is continually injected at the $x_1 = 0$ boundary. This data includes the measurements of conductivity $K_i^*:= K(\vec{x}_i^K)$, hydraulic head $h_i^*:= h(\vec{x}_i^h)$, and concentration $C_i^*:= C(\vec{x}_i^C)$ at the locations $\{ \vec{x}_i^K \} _{i=1}^{N_K}$, $\{ \vec{x}_i^h \} _{i=1}^{N_h}$, and $\{ \vec{x}_i^C \} _{i=1}^{N_C}$, respectively, where  $N_K$,  $N_h$,  and $N_C$ are the number of measurements of each variable. 
 Our objective is to learn the conductivity, head, and concentration fields based on these measurements. We further assume that the concentration, hydraulic head, and conductivity data can be accurately modeled by the steady-state Darcy flow: 
\begin{equation}\label{eq:diffusion}
\left\{
	\begin{array}{ll}
	\begin{split}
	         \vec{v}(\vec{x}) & = -\frac{K(\vec{x})}{\phi} \nabla h(\vec{x}) \\
		\nabla \cdot  \vec{v}(\vec{x}) & = 0, \quad \vec{x} \in \Omega \\
		h(\vec{x}) & = {H_2}, \quad {x_1} = {L_1} \\
		-K(\vec{x}) \partial h(\vec{x}) /  \partial x_1 & = q, \quad x_1 = 0 \\
		-K(\vec{x}) \partial h(\vec{x}) /  \partial x_2 & = 0, \quad x_2 = 0 \:{\rm{or}}\: x_2 = L_2
	\end{split}
	\end{array} \right.
\end{equation}
and advection--dispersion equation: 
\begin{equation}\label{eq:advdis}
\left\{
\begin{array}{ll}
	\begin{split}
		\nabla \cdot [{\vec{v}} (\vec{x}) C(\vec{x})] & = \nabla \cdot [ \vec{D} \nabla C(\vec{x})], \quad \vec{x} \in \Omega \\
		C(\vec{x}) & = C_0(x_2), \quad x_1 = 0 \\
   		\partial C(\vec{x}) /  \partial x_1 & = 0, \quad x_1 = L_1 \\
    	\partial C(\vec{x}) /  \partial x_2 & = 0, \quad x_2 = 0 \quad{\rm{or}}\quad x_2 = L_2
    \end{split}
\end{array} \right.
\end{equation}
where $\phi$ is the effective porosity of the medium, $\vec{v}$ is the average pore velocity, and $\vec{D}$ is the dispersion coefficient:
\begin{equation}
	\vec{D} = D_w \tau \vec{I} + \vec{\alpha} {||\vec{v}||}_2.
\end{equation}
Here, $\vec{I}$ is the identity tensor, $D_w$ is the diffusion coefficient, $\tau$ is the tortuosity of the medium, and $\vec{\alpha}$ is the dispersivity tensor with the diagonal components $\alpha_{L}$ and $\alpha_{T}$.
The conductivity  $K(\vec{x})$ is assumed to be unknown except at the measurement locations  $\{ \vec{x}_i^K \} _{i=1}^{N_K}$. 

In the following simulations, we set the parameters as: $L_1 = 1 \: \rm{m}$, $L_2 = 0.5 \: \rm{m}$, $H_2=0 \: \rm{m}$, $q=1 \: \rm{m/hr}$, $C_0(x_2)=c\exp(-\frac{(x_2 - L_2/2)^2}{\epsilon^2})$, $c=1 \: \rm{Kg/m^3}$, $\epsilon = 0.25 \: \rm{m}$ 
, $\phi=0.317$, $D_w=0.09 \: \rm{m^2/hr}$, $\tau = \phi^{1/3} = 0.681$, $\alpha_L = 0.01 \: \rm{m}$, and $\alpha_T = 0.001 \: \rm{m}$.

\begin{figure}[ht!]
	\centering
	\includegraphics[angle=0,width=4.6in]{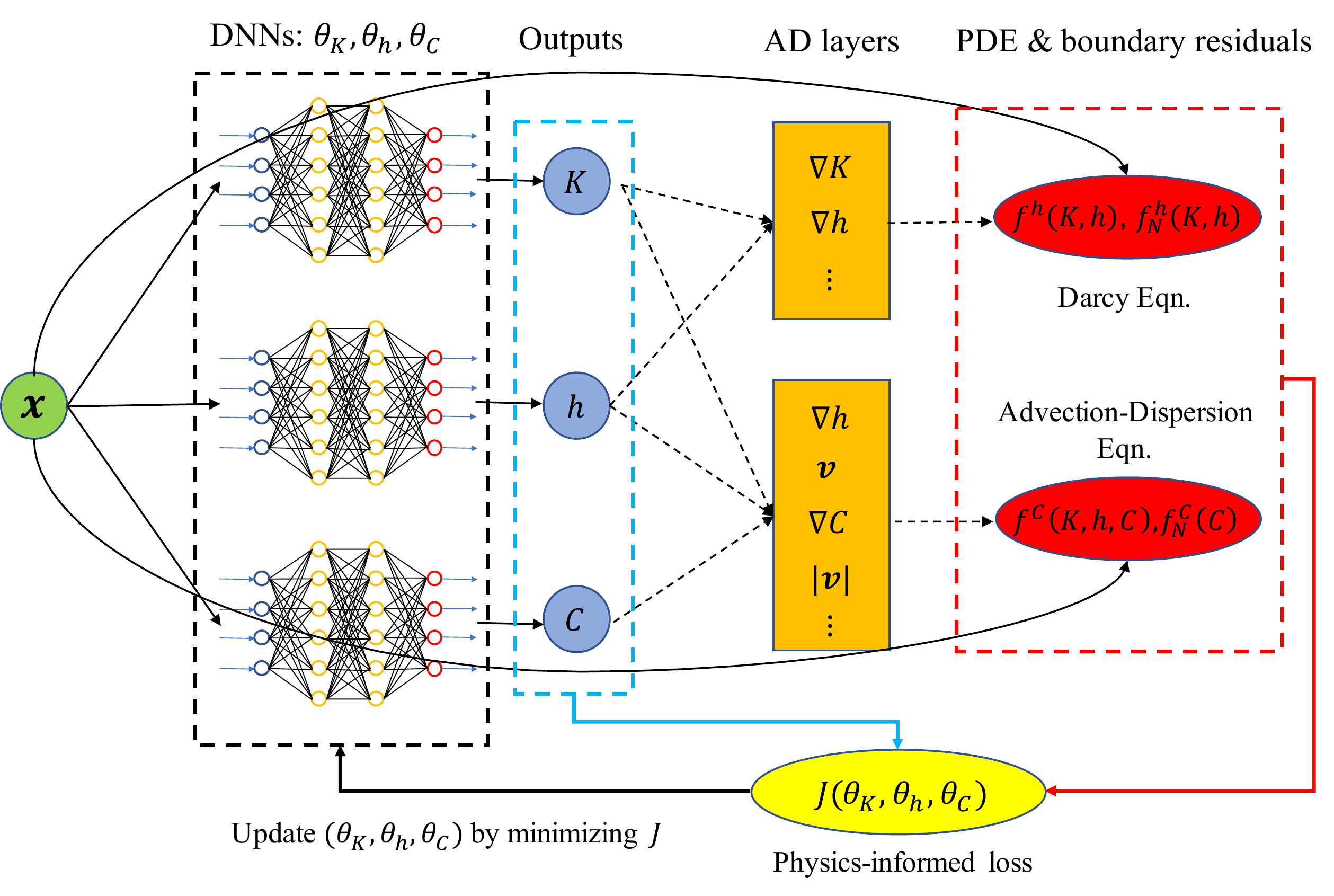}
	\caption{A schematic diagram of the PINN method for multiphysics data assimilation in subsurface transport problems. Three DNNs are used to represent the unknown $K(\vec{x})$, $h(\vec{x})$, and $C(\vec{x})$ fields. Spatial derivatives of these fields in the PDE and boundary condition residuals are computed with AD. The DNNs are trained by minimizing the ``physics-informed'' loss function.}
	\label{fig:sket_MPINN}
\end{figure}

We start by defining the DNN representations of $K(\vec{x})$, $h(\vec{x})$, and $C(\vec{x})$ as:
\begin{equation}
\begin{split}
	\hat{K}(\vec{x})  &:= \mathcal{N}_K (\vec{x};\theta_K) \\ 
	\hat{h}(\vec{x})  &:= \mathcal{N}_h (\vec{x};\theta_h) \\
	\hat{C}(\vec{x})  &:= \mathcal{N}_C (\vec{x};\theta_C)
\end{split}
\end{equation}
where $\theta_K$, $\theta_h$, and $\theta_C$ are the vectors of parameters associated with each neural network. Next, the residuals of equations \eqref{eq:diffusion} and \eqref{eq:advdis} are expressed in terms of  $\theta_K$, $\theta_h$, and $\theta_C$ as:
\begin{subequations} \label{eq:residual_pde}
	\begin{align}
	f^h(\vec{x};\theta_K,\theta_h) & =  \nabla \cdot [\hat{K}(\vec{x};\theta_K) \nabla \hat{h}(\vec{x},\theta_h)] \\
	f^C(\vec{x};\theta_K,\theta_h,\theta_C) & = -\frac{1}{\phi} \hat{K}(\vec{x};\theta_K) \nabla \hat{h}(\vec{x},\theta_h) \cdot \nabla \hat{C}(\vec{x},\theta_C) - \nabla \cdot [ \vec{D} \nabla \hat{C}(\vec{x},\theta_C)].
	\end{align}
\end{subequations}
To enforce the Neumann boundary conditions for equations (\ref{eq:diffusion}) and (\ref{eq:advdis}), we define DNNs approximating fluxes at the boundaries:
\begin{equation}\label{eq:residual_neum_h}
\begin{split}
f^h_{N1}(\vec{x};\theta_K,\theta_h) & = - \hat{K} (\vec{x}) \partial \hat {h} (\vec{x}) / \partial x_1 - q, \\
f^h_{N2}(\vec{x};\theta_K,\theta_h) & = - \hat{K} (\vec{x}) \partial \hat {h} (\vec{x}) / \partial x_2,
\end{split}
\end{equation}
and
\begin{align}  \label{eq:residual_neum_c}
\begin{split} % only use one notation for Eq.
		f^C_{N1}(\vec{x};\theta_C) & =  \partial \hat {C} (\vec{x}) / \partial x_1, \\
		f^C_{N2}(\vec{x};\theta_C) & = \partial \hat {C} (\vec{x}) / \partial x_2.
\end{split}
\end{align}
The  loss function is then defined as:
\begin{equation}\label{eq:loss_mpinn}
\begin{split}
	J(\theta_K,\theta_h,\theta_C) & = J_d (\theta_K,\theta_h,\theta_C) + J_f^h(\theta_K,\theta_h) + J_f^C(\theta_K,\theta_h,\theta_C) \\
				  & \quad + J_{N1}^h (\theta_K,\theta_h) + J_{N2}^h (\theta_K,\theta_h) + J_{N1}^C (\theta_C) + J_{N2}^C (\theta_C) \\
				  & \quad + J_b^h (\theta_h) + J_b^C (\theta_C),
\end{split}
\end{equation}
where the loss due to mismatch with data is
\begin{eqnarray} 
J_d(\theta_K,\theta_h,\theta_C)  
& = 
\frac{1}{N_K} \sum_{i}^{N_K} [\hat{K} (\vec{x}_i^K; \theta_K) - K_i^*]^2 \\ \nonumber
& +  
\frac{1}{N_h} \sum_{i}^{N_h} [\hat{h} (\vec{x}_i^h; \theta_h) - h_i^*]^2 \\ \nonumber
& +
\frac{1}{N_C} \sum_{i}^{N_C} [\hat{C} (\vec{x}_i^C; \theta_C) - C_i^*]^2,
\end{eqnarray}
and the losses due to PDE constraints and boundary conditions are: 
\begin{eqnarray} \label{physics_loss}
% PDE loss
J_f^h(\theta_K,\theta_h) & = \frac{1}{|\mathcal{T}_f^h|} \sum_{\vec{x} \in \mathcal{T}_f^h} [f^h(\vec{x};\theta_K,\theta_h) ]^2, \\ \nonumber
J_f^C(\theta_K,\theta_h,\theta_C) & = \frac{1}{|\mathcal{T}_f^C|} \sum_{\vec{x} \in \mathcal{T}_f^C} [ f^C(\vec{x};\theta_K,\theta_h,\theta_C) ]^2, \\ \nonumber
% Neumann loss
J_{N1}^h (\theta_K,\theta_h) &= \frac{1}{|\mathcal{T}_{N1}^h|} \sum_{\vec{x} \in \mathcal{T}_{N1}^h} [f^h_{N1}(\vec{x};\theta_K,\theta_h) ]^2, \\ \nonumber
J_{N2}^h (\theta_K,\theta_h) &= \frac{1}{|\mathcal{T}_{N2}^h|} \sum_{\vec{x} \in \mathcal{T}_{N2}^h} [f^h_{N2}(\vec{x};\theta_K,\theta_h) ]^2, \\ \nonumber
J_{N1}^C (\theta_C) &= \frac{1}{|\mathcal{T}_{N1}^C|} \sum_{\vec{x} \in \mathcal{T}_{N1}^C} [f^C_{N1}(\vec{x};\theta_C)]^2, \\ \nonumber
J_{N2}^C (\theta_C) &= \frac{1}{|\mathcal{T}_{N2}^C|} \sum_{\vec{x} \in \mathcal{T}_{N2}^C} [f^C_{N2}(\vec{x};\theta_C)]^2,\\ \nonumber
J_b^h (\theta_h) & = \frac{1}{|\mathcal{T}_{b}^h|} \sum_{\vec{x} \in \mathcal{T}_{b}^h} [\hat{h} (\vec{x}; \theta_h) - h^*(\vec{x})]^2, \\ \nonumber
J_b^C (\theta_C) & = \frac{1}{|\mathcal{T}_{b}^C|} \sum_{\vec{x} \in \mathcal{T}_{b}^C} [\hat{C} (\vec{x}; \theta_C) - C^*(\vec{x})]^2.
\end{eqnarray}
In Eq. (\ref{physics_loss}), PDEs  \eqref{eq:diffusion} and \eqref{eq:advdis} are enforced at the residual points given by the sets $\mathcal{T}_f^h$ and $\mathcal{T}_f^C$, respectively, where $|\mathcal{T}_f^h| = N_f^h$ and $|\mathcal{T}_f^C| = N_f^C$. The terms with the subscripts $N_1$ or $N_2$ enforce the Neumann boundary conditions, and those with the subscript $b$ enforce the Dirichlet boundary conditions. A schematic diagram of the proposed framework is shown in Figure \ref{fig:sket_MPINN}. 

In this work, we compare three approaches for training $\hat{K}$: jointly training the DNNs $\hat{K}$,  $\hat{h}$, and $\hat{C}$ by minimizing the loss function (\ref{eq:loss_mpinn}) that we refer to as ``multiphysics-informed neural networks'' (MPINN); jointly training $\hat{K}$,  $\hat{h}$ by only enforcing the Darcy equation and boundary conditions  (\ref{eq:diffusion}) that we refer to as ``PINN-Darcy;'' and separately training  $\hat{K}$,  $\hat{h}$, and $\hat{C}$ using only data that we refer to as ``data-driven DNN.'' The performance of these three approaches is investigated and compared in sections \ref{sec:Result} and \ref{sec:diss}.

Considering that the loss function is highly nonlinear and non-convex with respect to the network parameters $\theta_K$, $\theta_h$, and $\theta_C$, we use the gradient descent minimization algorithms, including the  Adam~\cite{Kingma2014}, and L-BFGS-B~\cite{Byrd1995} algorithms. As suggested in~\cite{Raissi2019,Tartakovsky2018,Berg2018,Le2011}, 
the L-BFGS-B, a quasi-Newtown method, shows superior performance with a better rate of convergence, lower gradient vanishing, and a lower computational cost for problems with a relatively small amount of training data. Therefore, in the following numerical examples, we primarily adopt the L-BFGS-B method with the default settings from Scipy~\cite{Jones_scipy2001} and the Xavier initialization scheme~\cite{Glorot2010a}. 

Adding physics constraints complicates the loss function landscape and makes the minimization process more challenging. We demonstrate that the parameter and state estimation could be improved by first pre-training $\hat{K}$ and $\hat{h}$ with the PINN-Darcy approach and then retraining $\hat{K}$ and $\hat{h}$ with $\hat{C}$ using the MPINN approach.  We denote this approach as the ``sequential training.'' The approach where all three DNNs are initialized and trained jointly we will refer to as the ``simultaneous training.''  
%
%%%%%%%%%%%%%%%%%%%%%%%%%%%%%%%%%%%%%%%%%%%%%%%%%%%%
%%%     Numerical Examples     %%%
%%%%%%%%%%%%%%%%%%%%%%%%%%%%%%%%%%%%%%%%%%%%%%%%%%%%
\section{Numerical example 1: periodic conductivity field}\label{sec:Result}

\begin{figure}[htb]
	\centering
	\subfloat[$K$] {\includegraphics[angle=0,width=2.2in]{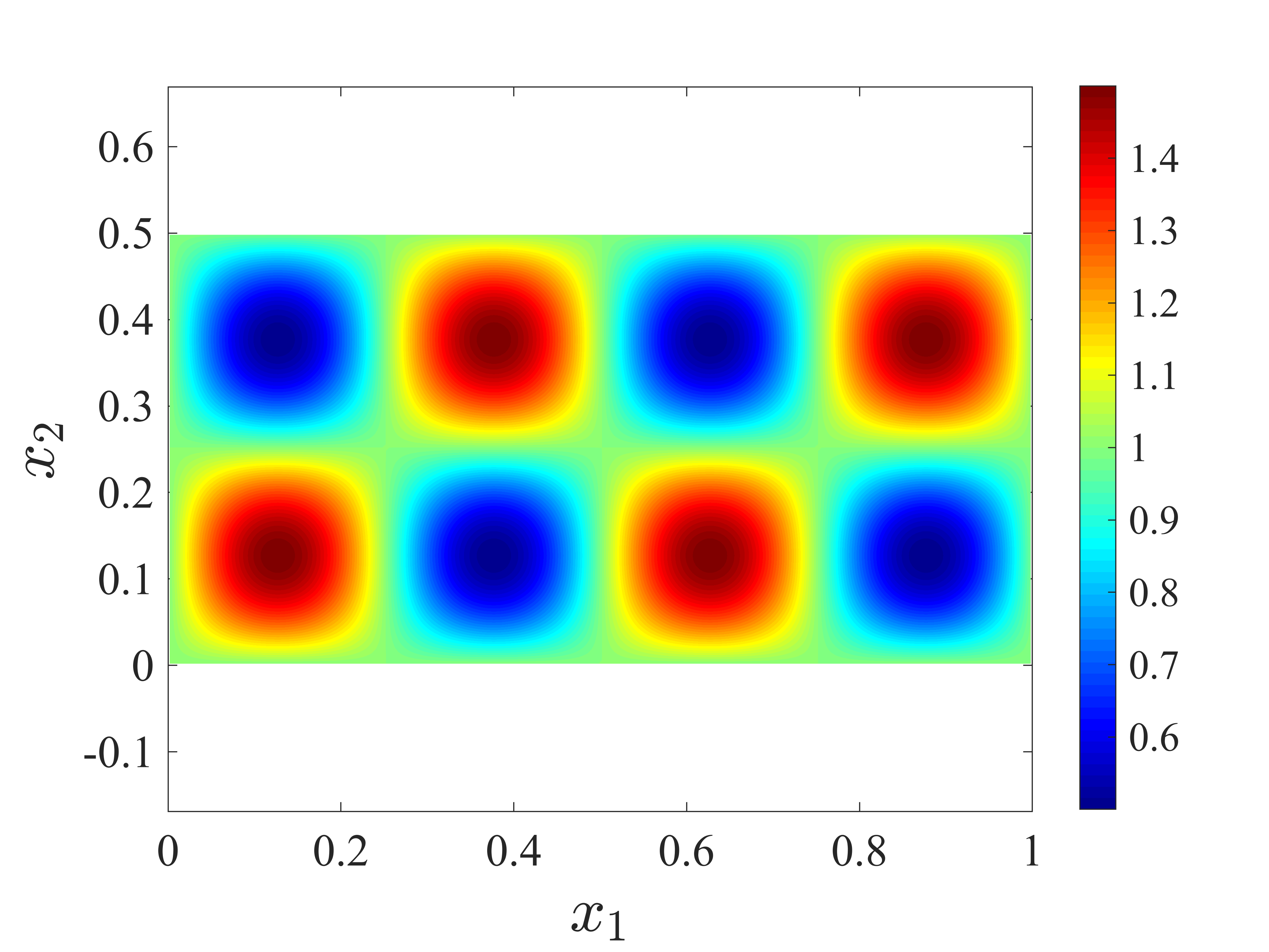}}
	\subfloat[$h$] {\includegraphics[angle=0,width=2.2in]{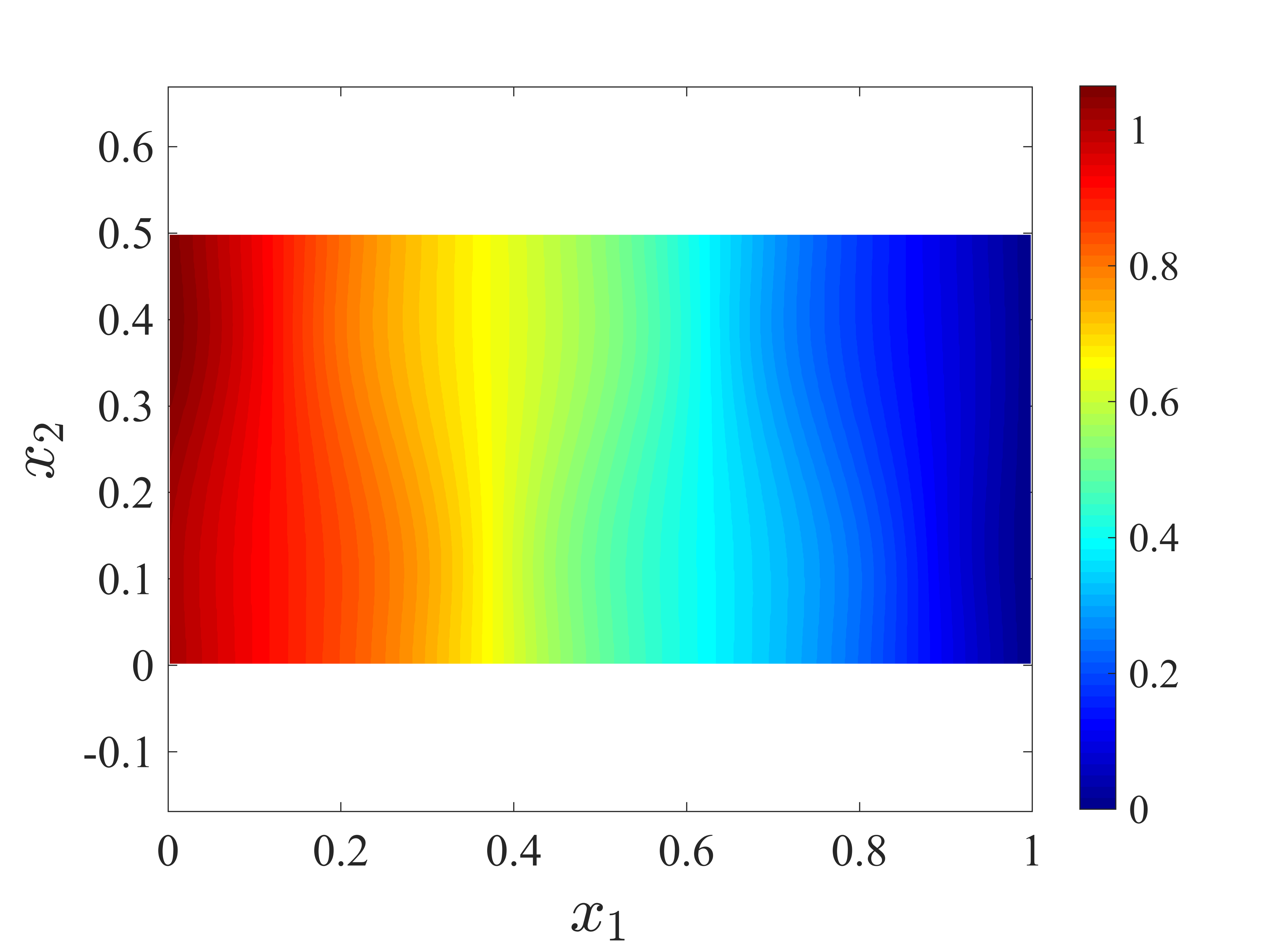}}
	
	\subfloat[$C$] {\includegraphics[angle=0,width=2.2in]{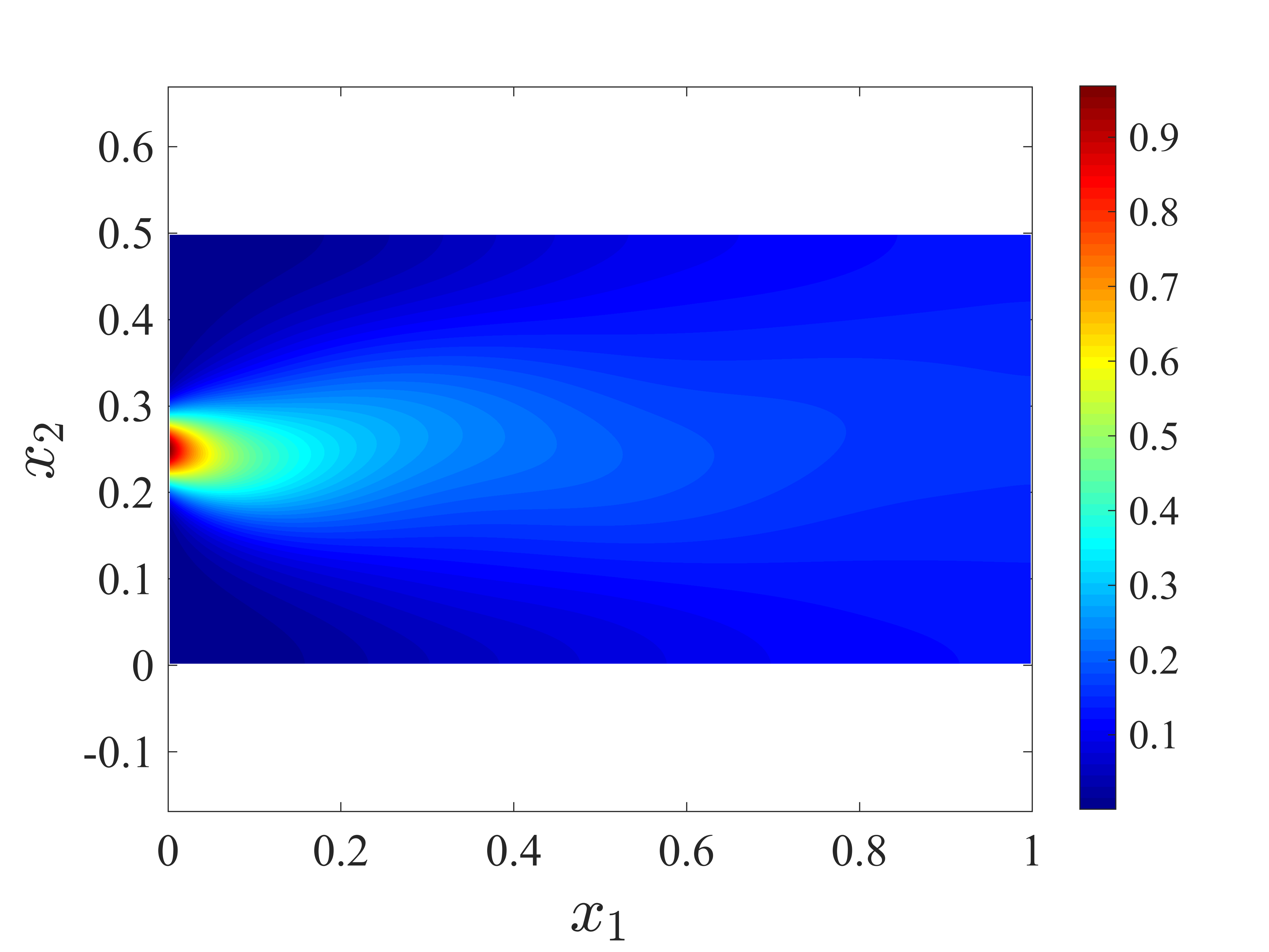}}
	\caption{Reference fields: (a) Conductivity $K$, (b) hydraulic head $h$, and (c) concentration $C$.}
	\label{fig:reference_sin_smooth}
\end{figure}

In this study, we test the proposed physics-informed neural network methods on synthetic data obtained by sampling numerical solutions of equations (\ref{eq:diffusion})-(\ref{eq:advdis}). These equations are solved using the finite-volume Subsurface Transport Over Multiple Phase (STOMP) code~\cite{White1995} with $256 \times 128$ cells. We obtain numerical solutions for several conductivity fields and refer to these conductivity fields and the corresponding numerical solutions as reference fields and use them to test the accuracy of the proposed methods. 

We first consider the reference conductivity field $K(\vec{x}) = 0.5 \sin(4 \pi x_1)\sin(4 \pi x_2) + 1$, shown in Figure \ref{fig:reference_sin_smooth}(a). Figures \ref{fig:reference_sin_smooth}(b)-(c) present the reference $h$ and $C$ fields.  In this case, $K(\vec{x})$ is a smooth periodic function, $h(\vec{x})$ is a smooth near-linear function, and  $C(\vec{x})$ is a highly nonlinear function  with sharp gradients near the solute injection point. To investigate the effect of different DNN architectures (i.e., the number of hidden layers and the number of neurons in each layer) on the performance of the physics-informed DNNs, we describe the DNN architecture as [-$m_1$-$m_2$-...-$m_{n_l}$-], where $n_l$ is the number of hidden layers and $m_i$ is the number of neurons in the $i$-th hidden layers.  All DNNs considered here  have a two-dimensional input layer (corresponding to $x_1$ and $x_2$) and a one-dimensional output layer (corresponding to scalar quantities $K$, $h$, or $C$).  In this section, we use the L-BFGS-B method with the Scipy default hyperparameters~\cite{Jones_scipy2001} to minimize the loss functions. 

We quantify the error between the estimated and reference fields in terms of the relative $L_2$ errors, defined as:
\begin{equation}  \label{eq:relative_err}
\epsilon^\gamma : = \frac{1}{\int_{\Omega} \gamma(\vec{x})^2 \rm{d} \vec{x}}{\int_{\Omega} [\gamma(\vec{x})-\hat{\gamma} (\vec{x})]^2 \rm{d} \vec{x}}, \quad \text{for} \: \gamma = K, h, C
\end{equation}
where $K(\vec{x})$, $h(\vec{x})$, and $C(\vec{x})$ are the reference fields.

%
%
%%% Regression %%%
\subsection{Pure data-driven DNNs for function approximation}\label{sec:pure_dnn}
\begin{figure}[htb]
	%\begin{figure} [ht!]
	\centering
	\subfloat[] {\includegraphics[angle=0,width=2.2in]{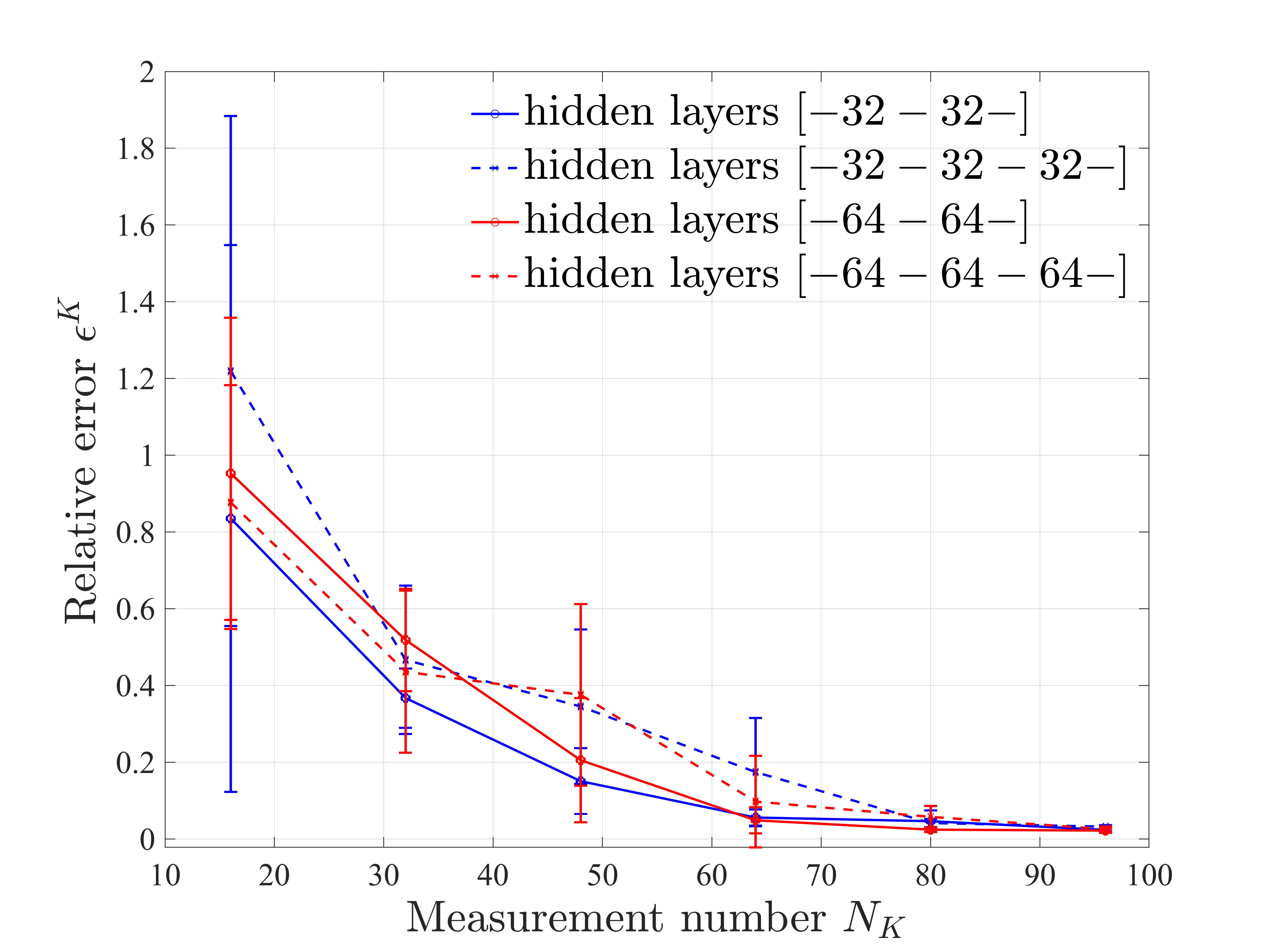}}
	\subfloat[] {\includegraphics[angle=0,width=2.2in]{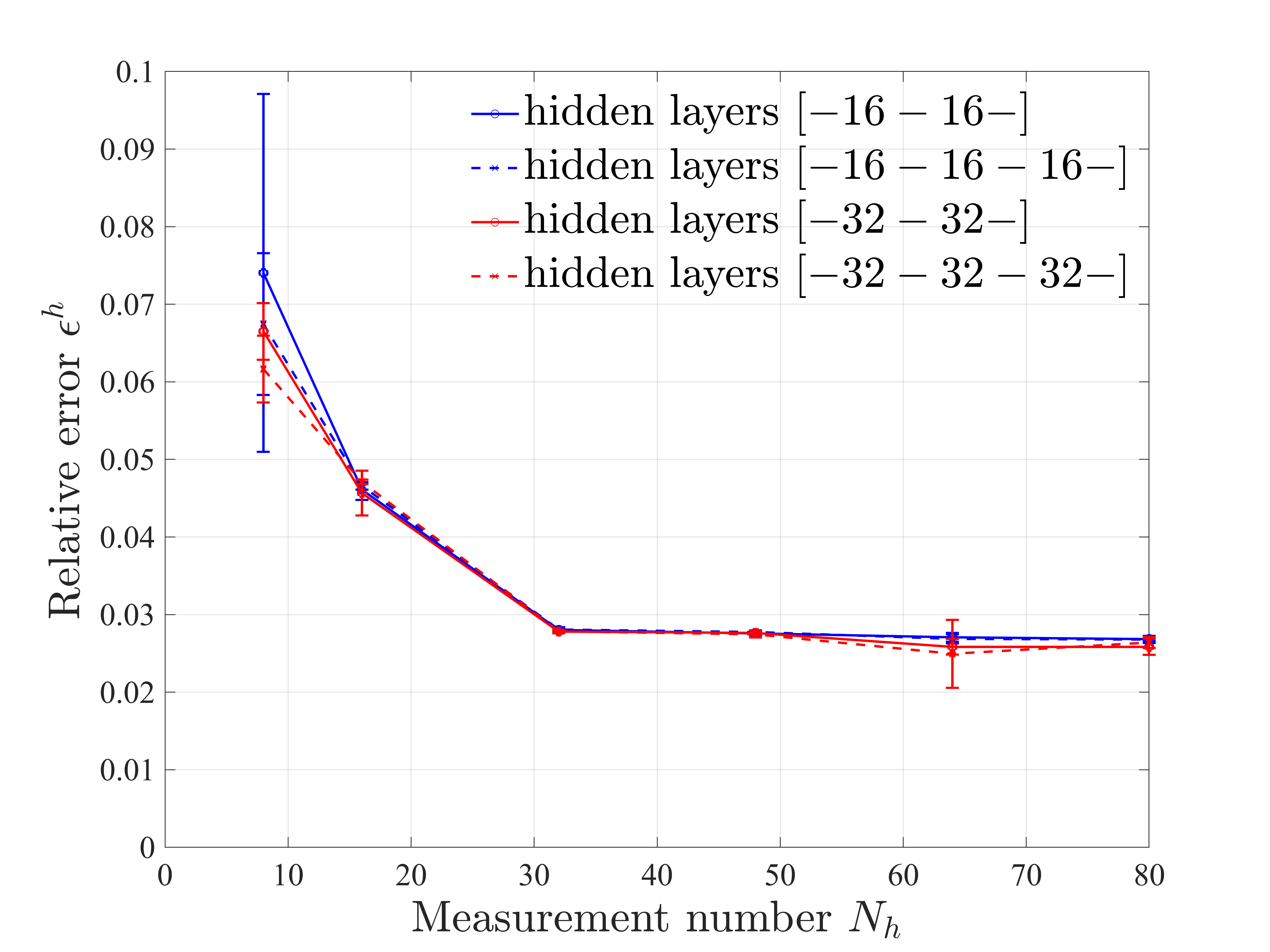}}
	
	\subfloat[] {\includegraphics[angle=0,width=2.2in]{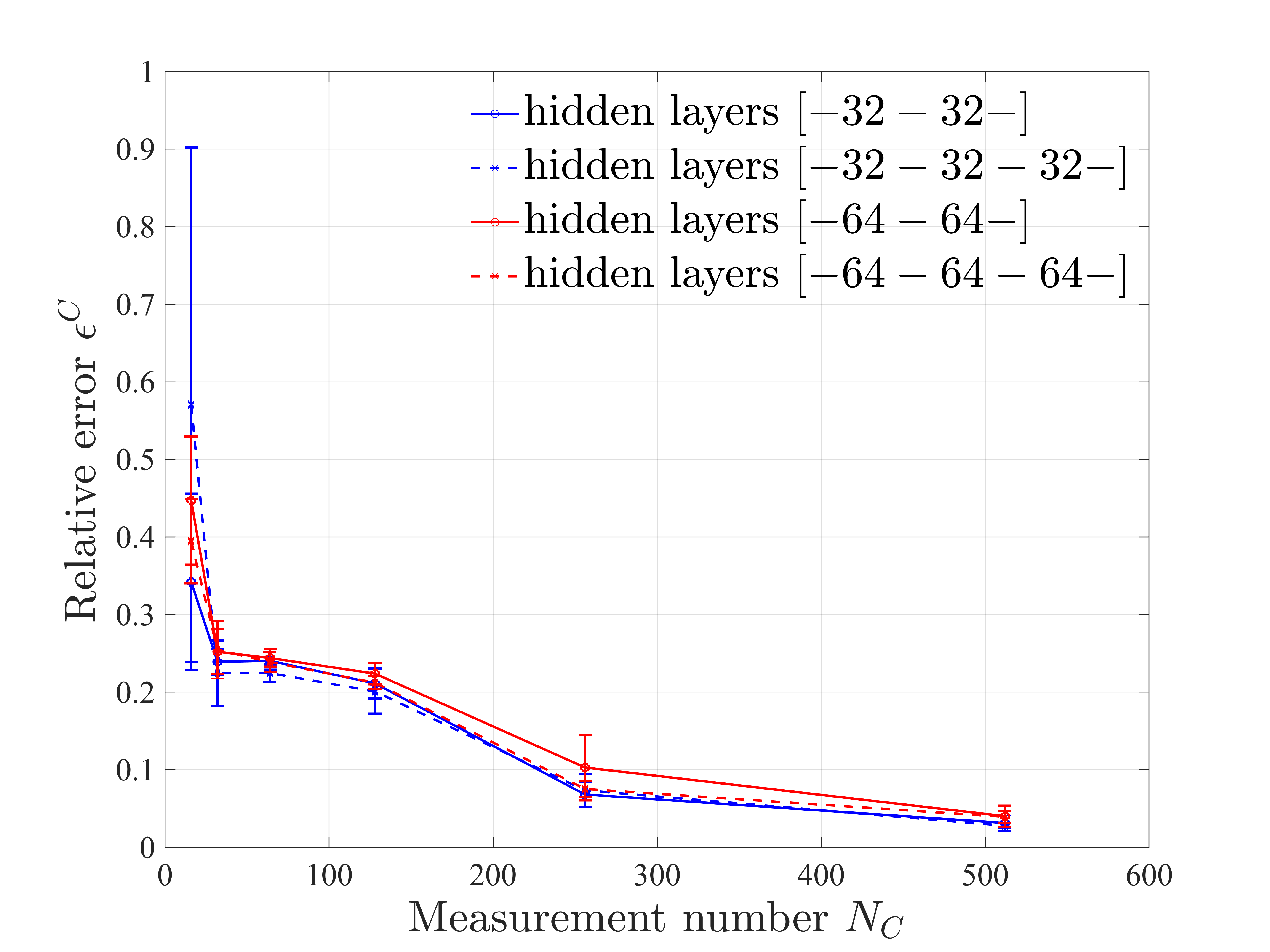}}
	\caption{The relative error versus the number of measurements  as a function of the DNN architecture for (a) conductivity $K(\vec{x})$, (b) hydraulic head $h(\vec{x})$, and (c) concentration $C(\vec{x})$ estimated with the data-driven DNNs. The mean and standard deviation are computed from five simulations corresponding to different DNN initializations.}
	\label{fig:regression}
\end{figure}

We first test the accuracy of the data-driven DNNs for learning $K$, $h$, and $C$ fields from their corresponding measurements. The DNN parameters $\theta_K$, $\theta_h$, and $\theta_C$ are found by solving a non-convex minimization problem and might depend on the initialization of the minimization algorithm. To estimate the effect of the DNN initialization on the error in the estimated parameters, for each variable ($K$, $h$, and $C$) we train DNNs with five different random initializations using Xavier's scheme and compute the mean error and standard deviation of the estimated variables. 

The average error and the error standard deviation in the estimated $K$, $h$, and $C$  fields are shown in Figure \ref{fig:regression} as a function of the number of observations for DNNs of different size. For all variables and network sizes, the error (both, the mean error and the standard deviation) decreases with an increasing number of measurements. For a larger number of measurements, the error does not significantly depend on the DNN size. On the other hand, the network size significantly affects the error for a smaller number of measurements. For the highly nonlinear $K$ and $C$ fields, figures \ref{fig:regression} (a) and (c) show that the smaller DNNs yield a better approximation than the larger ones. This is because larger DNNs need more data for accurate training. 
The $h$ field is very smooth and can be accurately approximated with a relatively small number of measurements regardless of the DNN size, as shown in Figure \ref{fig:regression} (b). Also, Figure \ref{fig:regression} clearly demonstrates that the number of measurements required to learn the field with a given accuracy strongly depends on the field's complexity.

For the smaller number of measurements, we observe a significant uncertainty (large standard deviation) due to the random DNN initialization in the data-driven models of $K$, $h$, and $C$.
This could be explained by the strong non-convexity of the loss function and the lack of regularization in the data-driven loss function. 
Several regularization techniques, including $L1$ and $L2$ regularizers, have been suggested for training DNNs with small data sets. One challenge with this approach is that the resulting DNNs strongly depend on the choice of regularizer. Instead, in the PINN-Darcy and MPINN methods, we use PDE constraints to improve the accuracy of DNN training with sparse data and demonstrate that these constraints act as a regularizer. The advantage of using physics constraints is that the resulting solutions satisfy the governing equations, while solutions obtained with the $L1$ or $L2$ regularization, in general, do not. 

\subsection{Data assimilation based on physics-informed neural networks}

\subsubsection{PINN-Darcy}\label{sec:PINN_Darcy}
\begin{figure}% [htb]
	\centering
	\subfloat[] {\includegraphics[angle=0,width=2.5in]{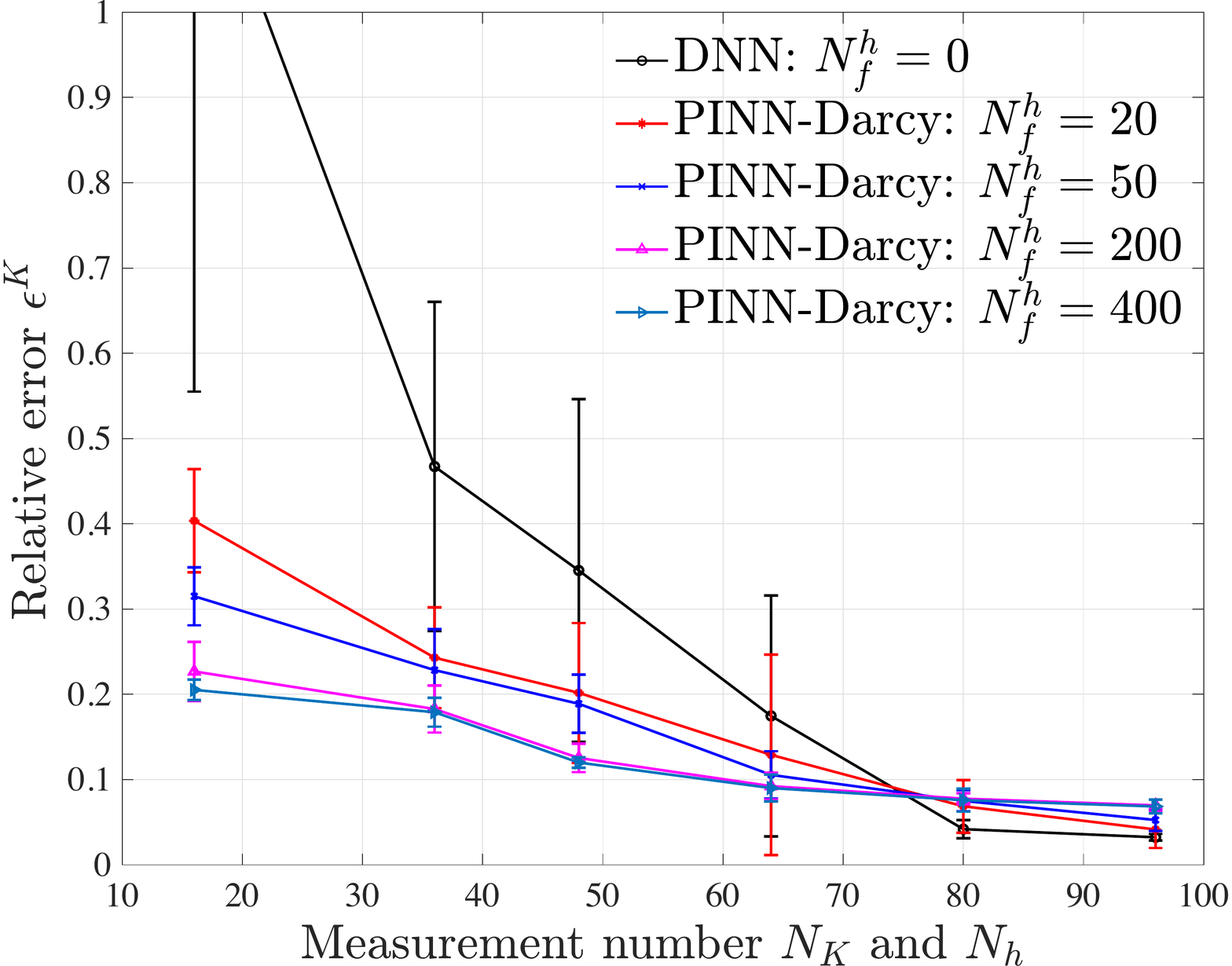}}
	\subfloat[] {\includegraphics[angle=0,width=2.5in]{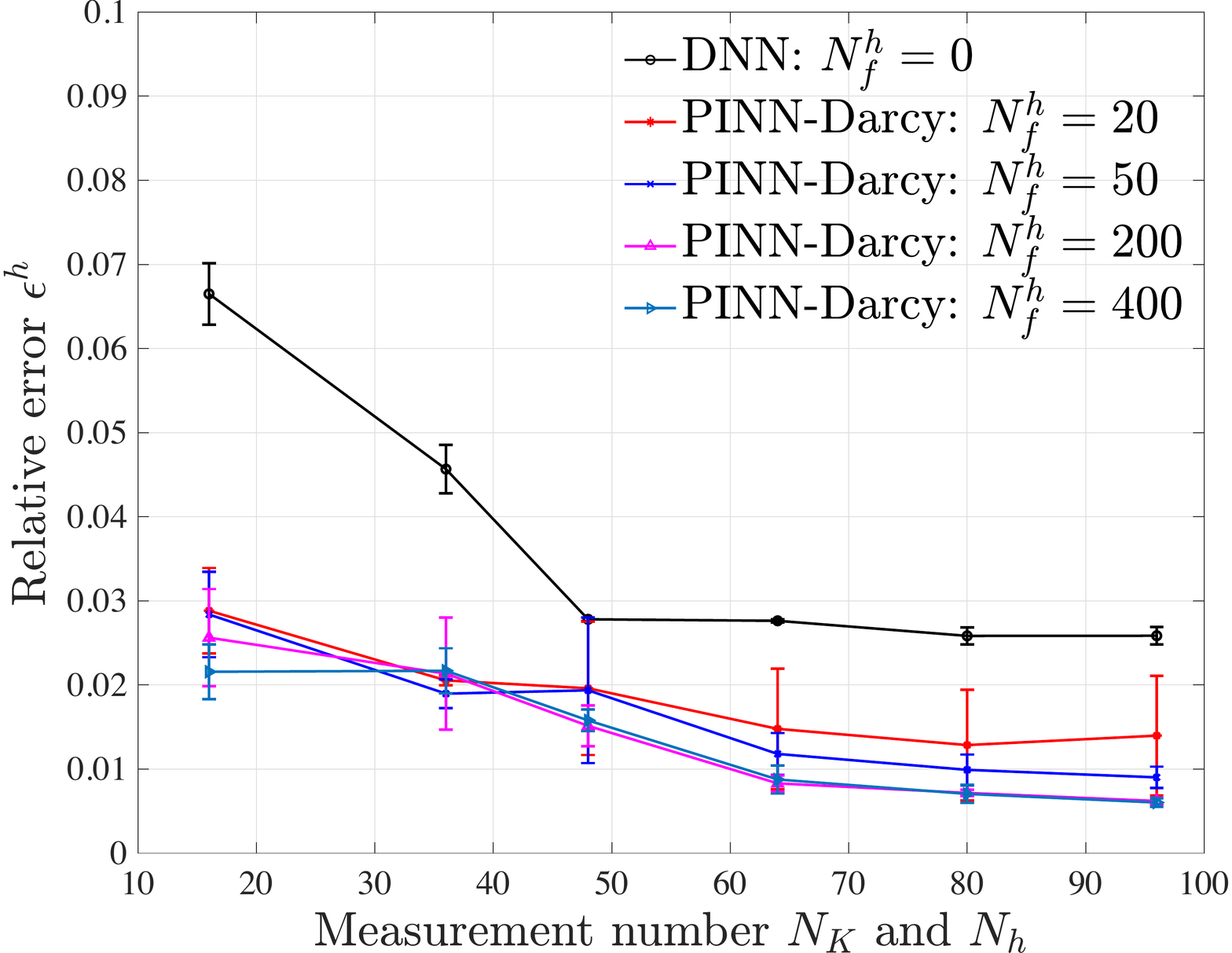}} \\
	\subfloat[] {\includegraphics[angle=0,width=2.5in]{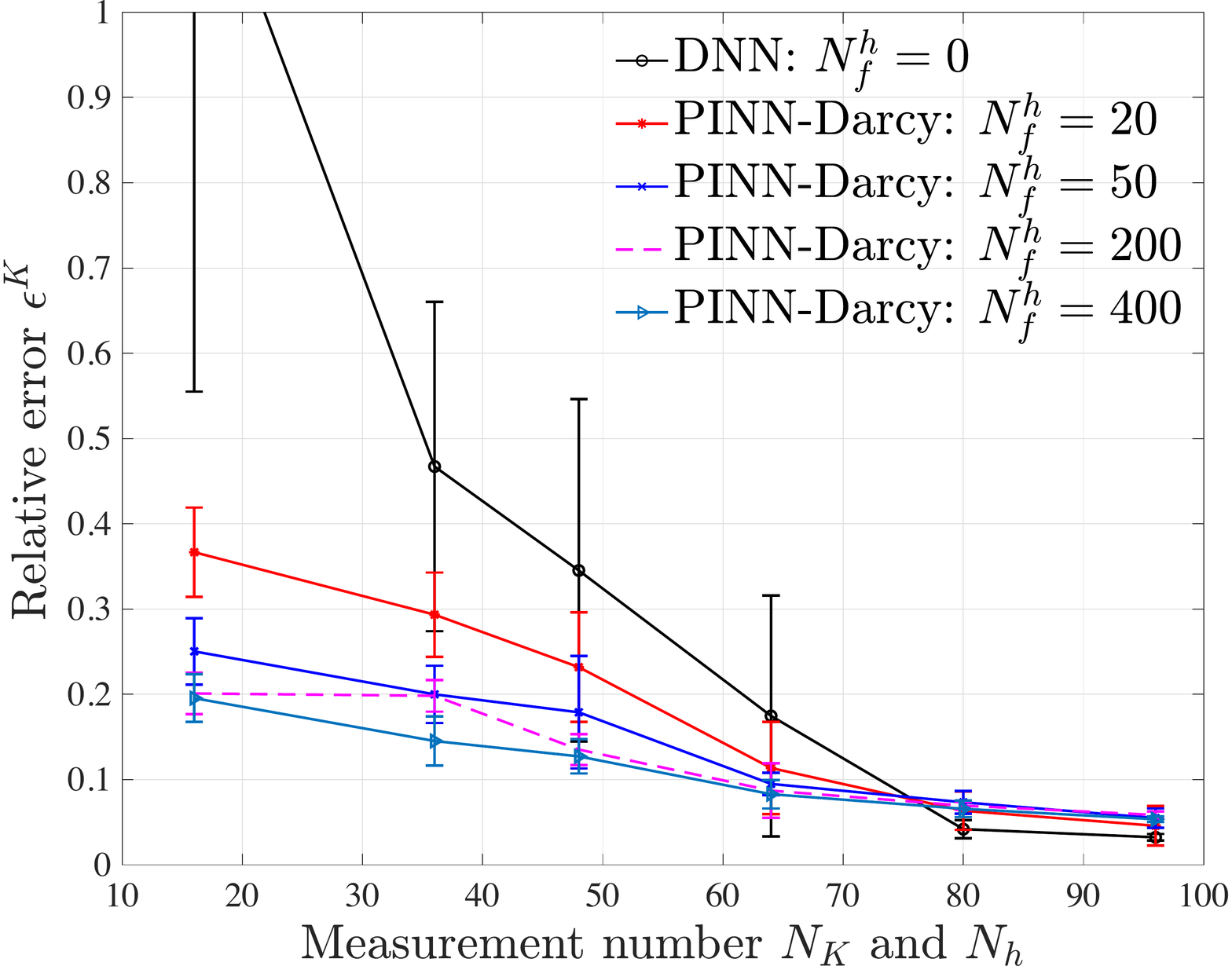}}
	\subfloat[] {\includegraphics[angle=0,width=2.5in]{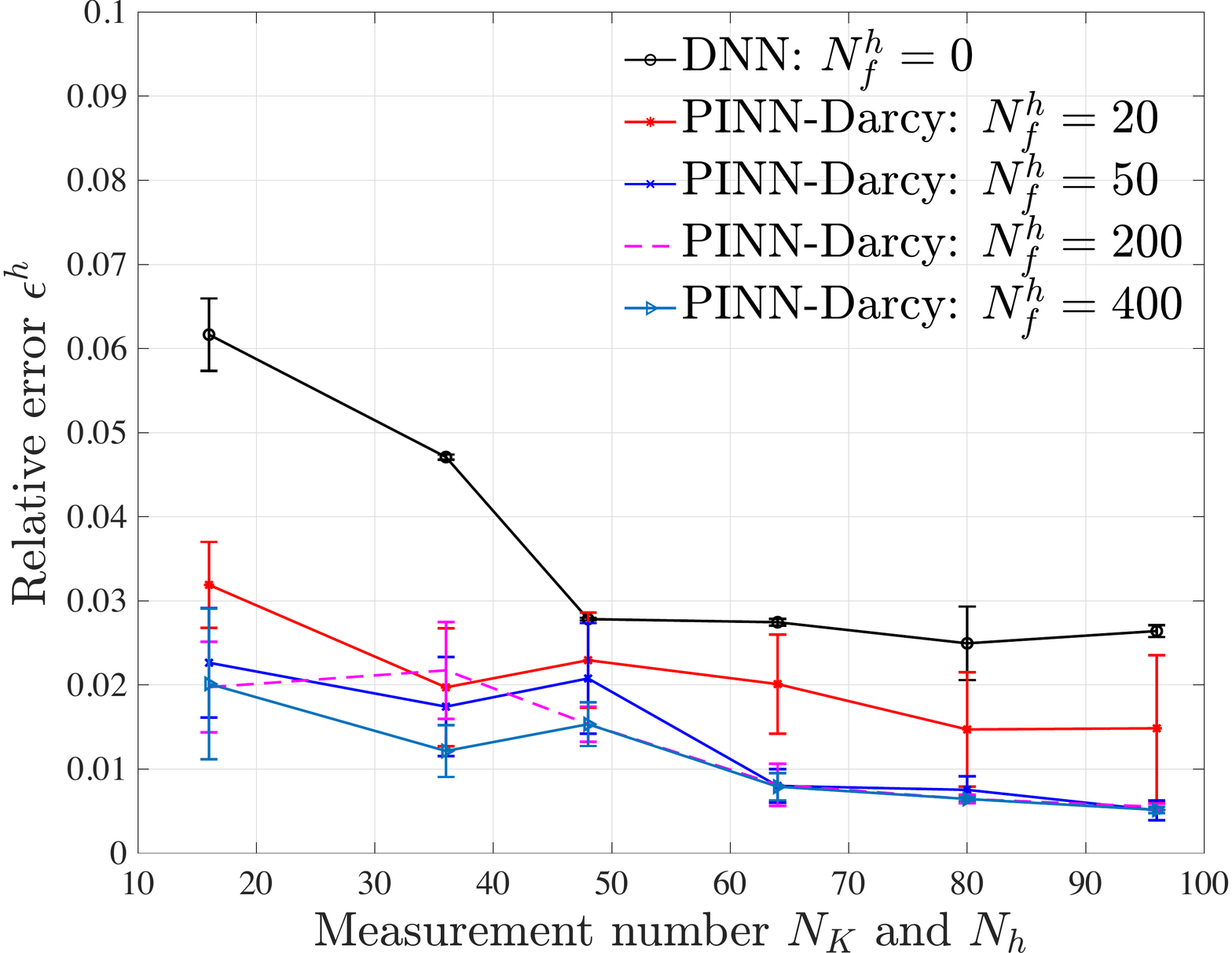}}
	\caption{The relative error versus number of measurements $N=N_K=N_h$ in the PINN-Darcy estimation of $K(\vec{x})$ (left column) and $h(\vec{x})$ (right column) as a function of the number of the residual points $N_f^h$. The top row corresponds to the simulations with smaller DNNs, including [-32-32-32-] for $\hat{K}$ and [-32-32-] for $\hat{h}$. The bottom row is obtained with larger DNNs, including  [-32-32-32-32-32-] for $\hat{K}$ and [-32-32-32-] for $\hat{h}$. The case with zero residual points ($N_f^h=0$) corresponds to the data-driven DNN method.}
	\label{fig:PINN-Darcy-conv}
\end{figure}

To demonstrate the proposed physics-informed framework for data assimilation, we first examine the PINN-Darcy approach, where the measurements of conductivity and hydraulic head and the Darcy equation \eqref{eq:diffusion} are used to jointly train the DNNs $\hat{K}$ and $\hat{h}$. The relative mean error and the standard deviation of the $K$ and $h$ fields  versus $N=N_K=N_h$ are plotted in Figure \ref{fig:PINN-Darcy-conv} for the different number of residual points $N_f^h$ and two different network sizes. For $N_f^h=0$, the PINN-Darcy method reduces to the data-driven DNN method. As before, for each case we train $\hat{K}$ and $\hat{h}$ five times with different initializations to compute the error's mean and standard deviation. In this test, the number and locations of measurements are randomly selected.

 Overall, the accuracy of $\hat{K}$ and $\hat{h}$ improves with increasing $N$ for all considered $N_f^h$. For $N<80$, the accuracy of  $\hat{K}$ increases with increasing $N_f^h$, i.e., both the mean error and the standard deviation decrease with an increasing number of residual points. The effect of enforcing the Darcy equation (i.e., having  $N_f^h>0$) is especially profound for ``sparse'' data.  For example, for $N=16$, the mean error in PINN-Darcy with ($N_f^h=400$) is $\approx 6$ times smaller than in the data-driven DNN. In addition, physics constraints allow training deeper and wider DNNs even with sparse data. 
The comparison of figures \ref{fig:PINN-Darcy-conv} (a) and (c) shows that the approximation error is slightly smaller for larger DNNs in the physics-informed DNN approach. We see the opposite trend in the data-driven DNN approach, as discussed in Section \ref{sec:pure_dnn}. 
For the large number of measurements (in this case, $N > 80$), the error decreases with deceasing $N_f^h$ and is smallest in the data-driven DNN approach. There are several reasons for the data-driven DNN approach to be more accurate than PINN-Darcy, including:  in this example, $N > 80$ measurements are sufficient to accurately train $\hat{K}$ without physics constraints, as evidenced by the small mean error and the error's standard deviation; the physics constraints in PINN-Darcy make the loss function more complicated and the loss function harder to minimize; and the physics model might not be exact. In our case, the synthetic data are sampled from a weak-form solution of the Darcy equation, while the (strong-form) PDE constraints are enforced in the loss function.  For the same number of measurements, the error (both, the mean error and the standard deviation) in the estimated $h$ is more than one order of magnitude smaller than in $K$ for both the data-driven DNN and PINN-Darcy methods. The small error in $h$ is due to the near-linear behavior of this field.   

%*** MPINN %%%

\begin{figure} %[htb]
	\centering
	\subfloat[] {\includegraphics[angle=0,width=2.5in]{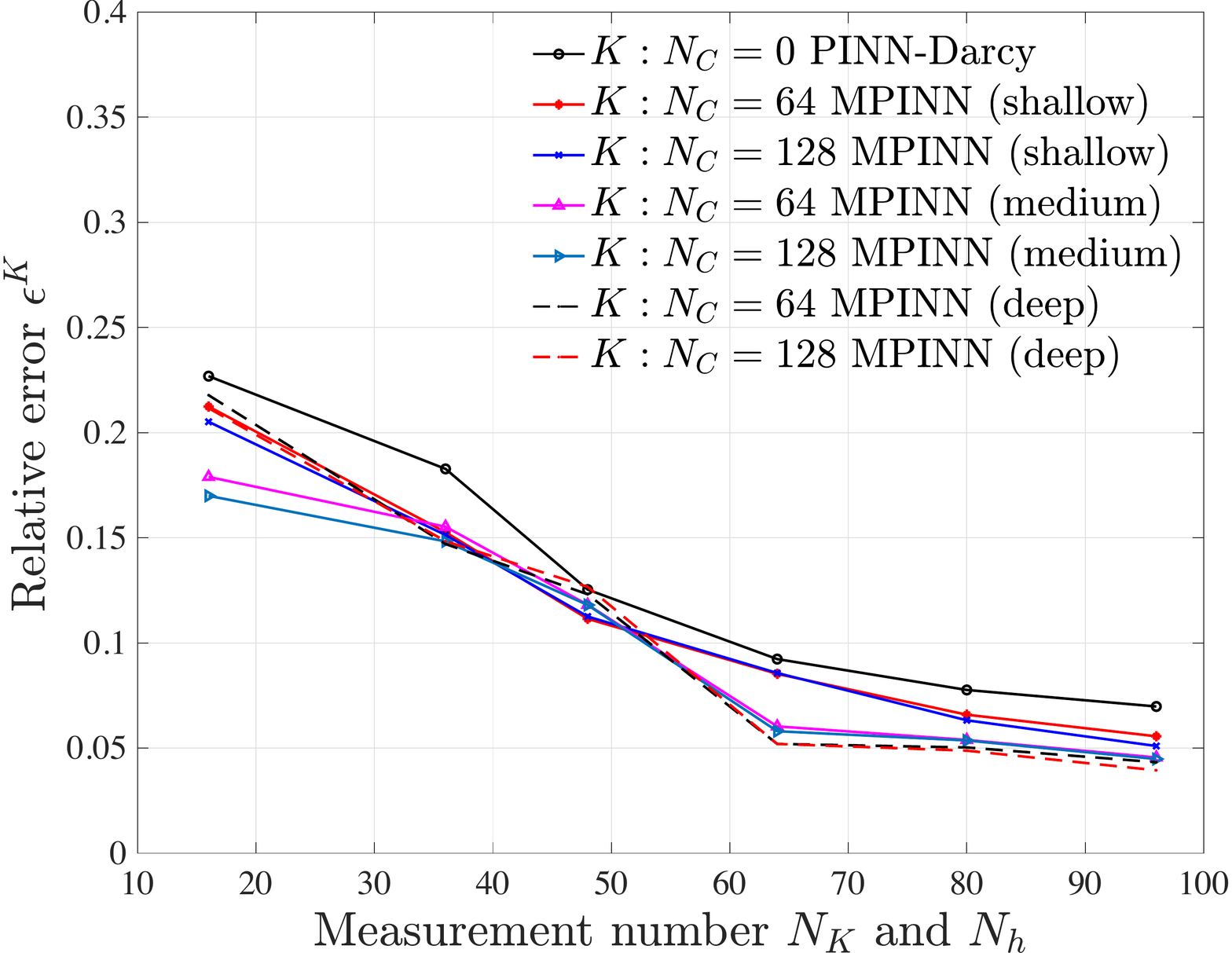}}\\
	\subfloat[] {\includegraphics[angle=0,width=2.5in]{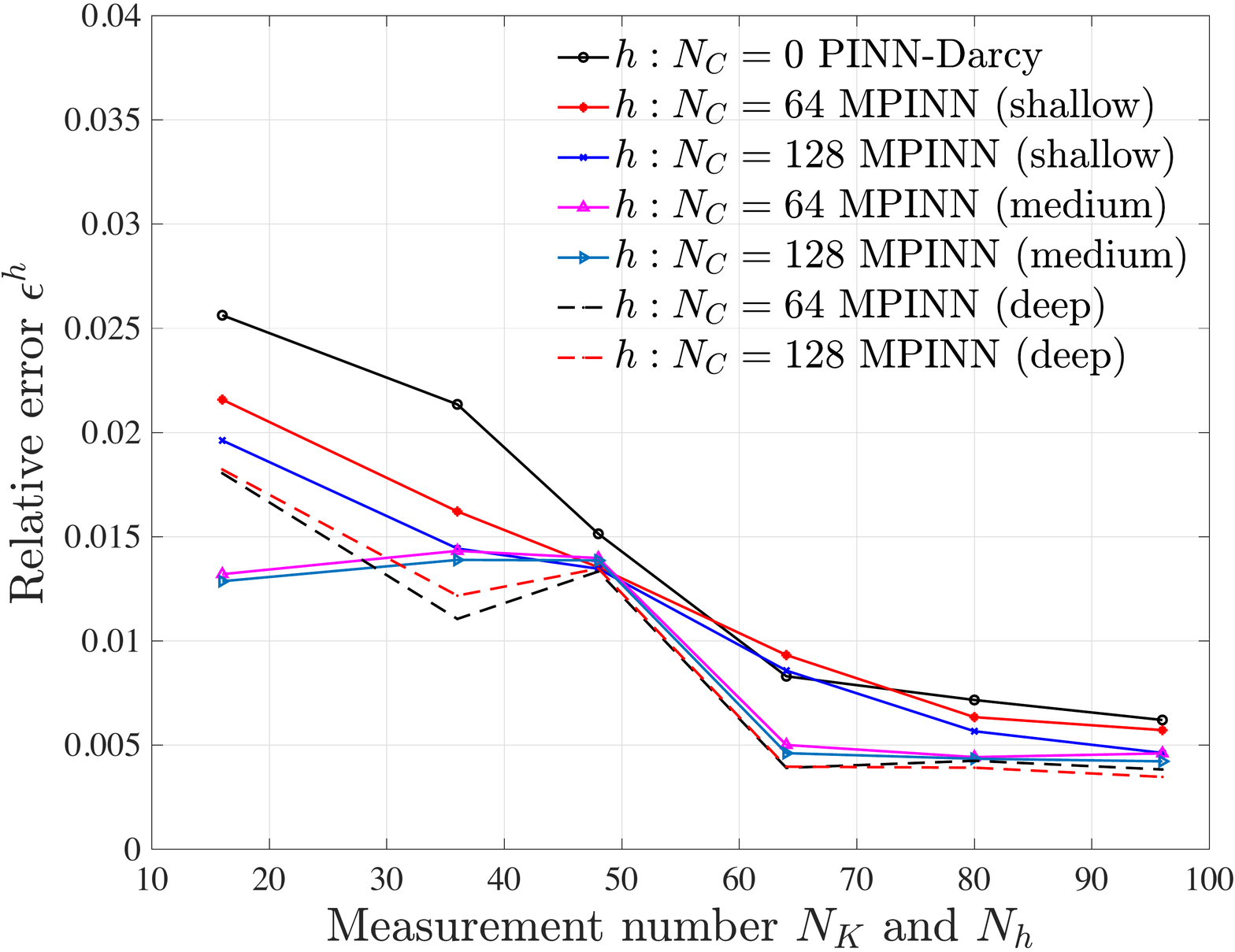}}
	\subfloat[] {\includegraphics[angle=0,width=2.5in]{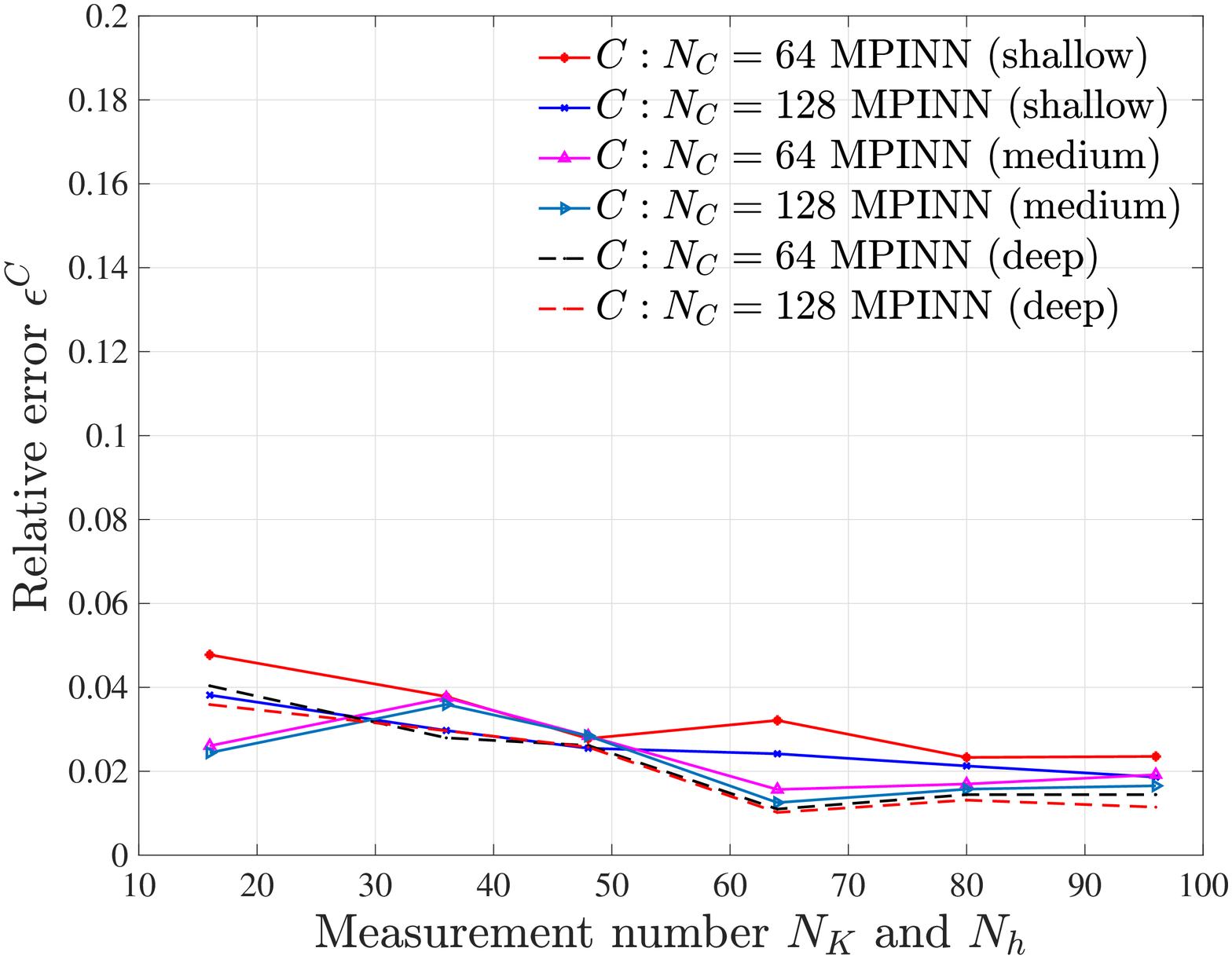}}
	\caption{The mean relative errors in the estimated (a)  $K(\vec{x})$, (b)  $h(\vec{x})$, and (c) $C(\vec{x})$ versus $N_K$ and $N_h$ using MPINN with $N_C =0, 64, 128$ and the ``shallow'' [-32,32,32-], ``medium'' [-32,32,32,32-], and ``deep'' [-32,32,32,32,32-]  DNNs $\hat{K}(\vec{x};\theta_K)$, $\hat{h}(\vec{x};\theta_h)$, and $\hat{C}(\vec{x};\theta_C)$. The other parameters are set to ($N_f^h = 200,N_f^C = 1000$). Sequential training is used in MPINN.  With $N_C =0$, MPINN is reduced to PINN-Darcy. }
	\label{fig:MPINN-conv}
\end{figure}

\subsubsection{MPINN}

In this section, we jointly train $\hat{K}(\vec{x};\theta_K)$, $\hat{h}(\vec{x};\theta_h)$, and $\hat{C}(\vec{x};\theta_C)$ for estimating $K$, $h$, and $C$ using the MPINN method with the proposed sequential training scheme. Figure \ref{fig:MPINN-conv} shows the mean error in the MPINN-estimated fields versus $N_K$, $N_h$, and $N_C$ for three different network sizes. The number of residual points is set to $N_f^h = 200$ and $N_f^C = 1000$. As before, our results show that the the estimation errors decrease with an increasing number of measurements. 
The MPINN method ($N_C>0$) improves the estimation of $K(\vec{x})$ and $h(\vec{x})$ fields relative to the PINN-Darcy method ($N_C=0$), demonstrating that the integration of $C$ measurements with $K$ and $h$ measurements by means of the PDE constraints improves the  $\hat{K}(\vec{x};\theta_K)$ and $\hat{h}(\vec{x};\theta_h)$ DNN training.  MPINN has the biggest impact on the $C$ field estimation, where the relative error reduces from more than 0.2 in the data-driven DNN (Figure \ref{fig:regression}(c)) to less than 0.05.   

Figure \ref{fig:MPINN-conv} also shows the effect of the DNN depth on the accuracy of the MPINN method. Specifically, we consider the ``shallow'' [-32,32,32-], ``medium'' [-32,32,32,32-], and ``deep'' [-32,32,32,32,32-] DNNs $\hat{K}(\vec{x};\theta_K)$, $\hat{h}(\vec{x};\theta_h)$, and $\hat{C}(\vec{x};\theta_C)$. The shallow DNNs perform the worst because of their limited representation ability. The medium-size DNNs perform the best with a smaller number of $K$ measurements, and the deep DNNs have the best performance with a larger number of measurements. This is because deeper DNNs are more representative but  need more data for training. We note that the PDE constraints allow deeper DNNs to be trained with sparse data; see also our results in Section \ref{sec:PINN_Darcy}.

%
%*** Sequential Training %%%
\begin{figure} %[htb]
	\centering
	\subfloat[] {\includegraphics[angle=0,width=2.5in]{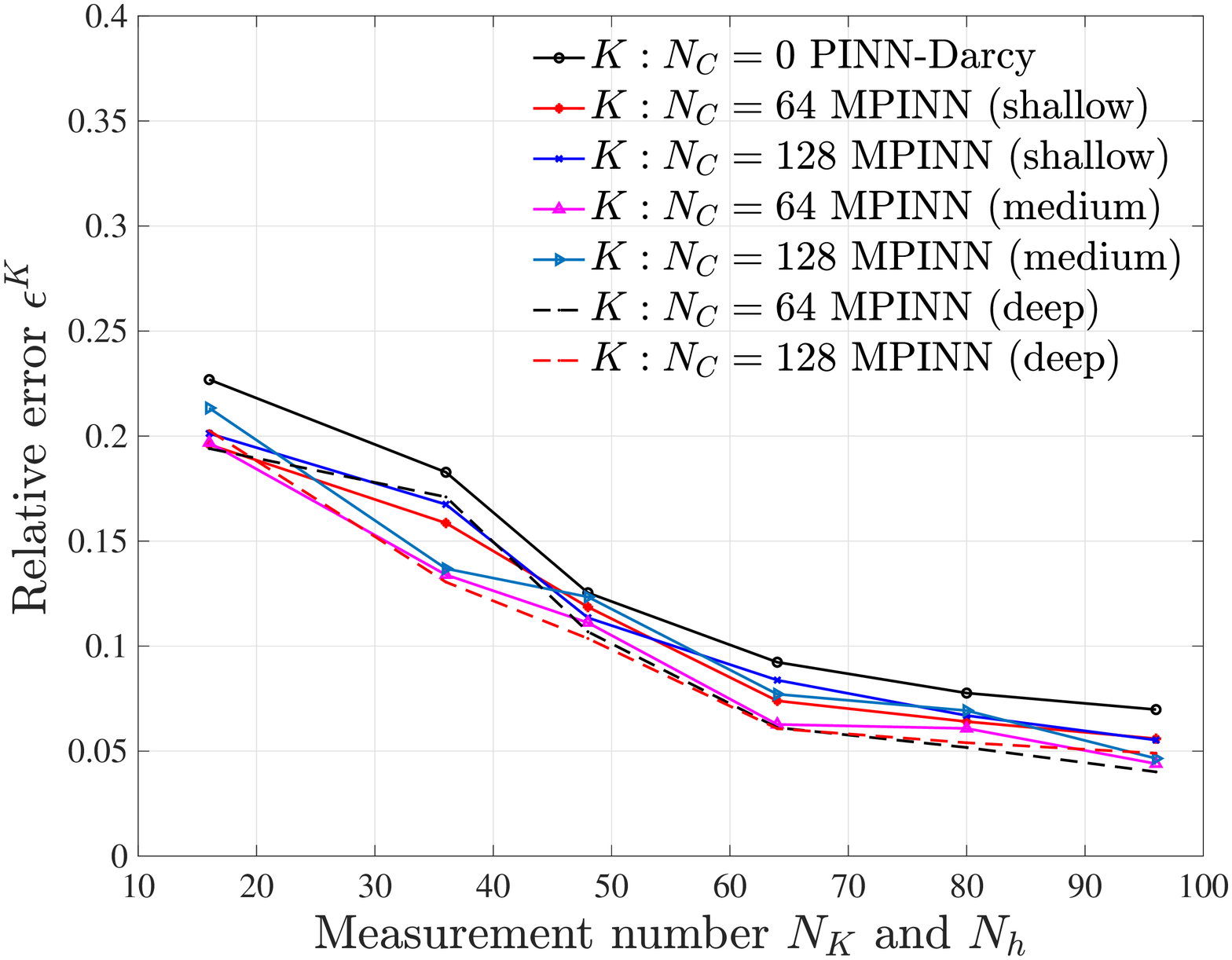}}\\
	%	\hspace*{\fill}
	%	\medskip # used for subfigure to separete vertically
	\subfloat[] {\includegraphics[angle=0,width=2.5in]{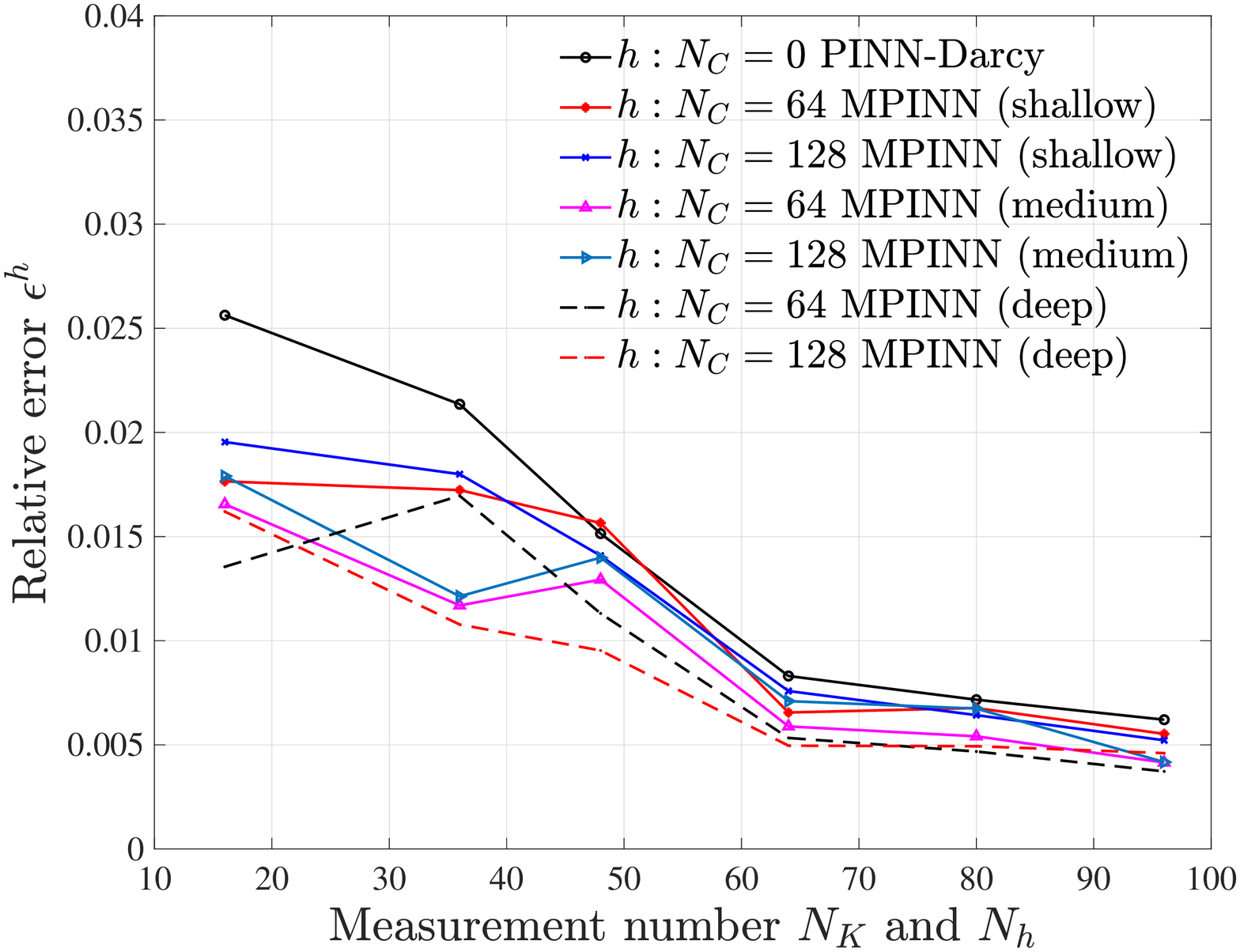}}
	\subfloat[] {\includegraphics[angle=0,width=2.5in]{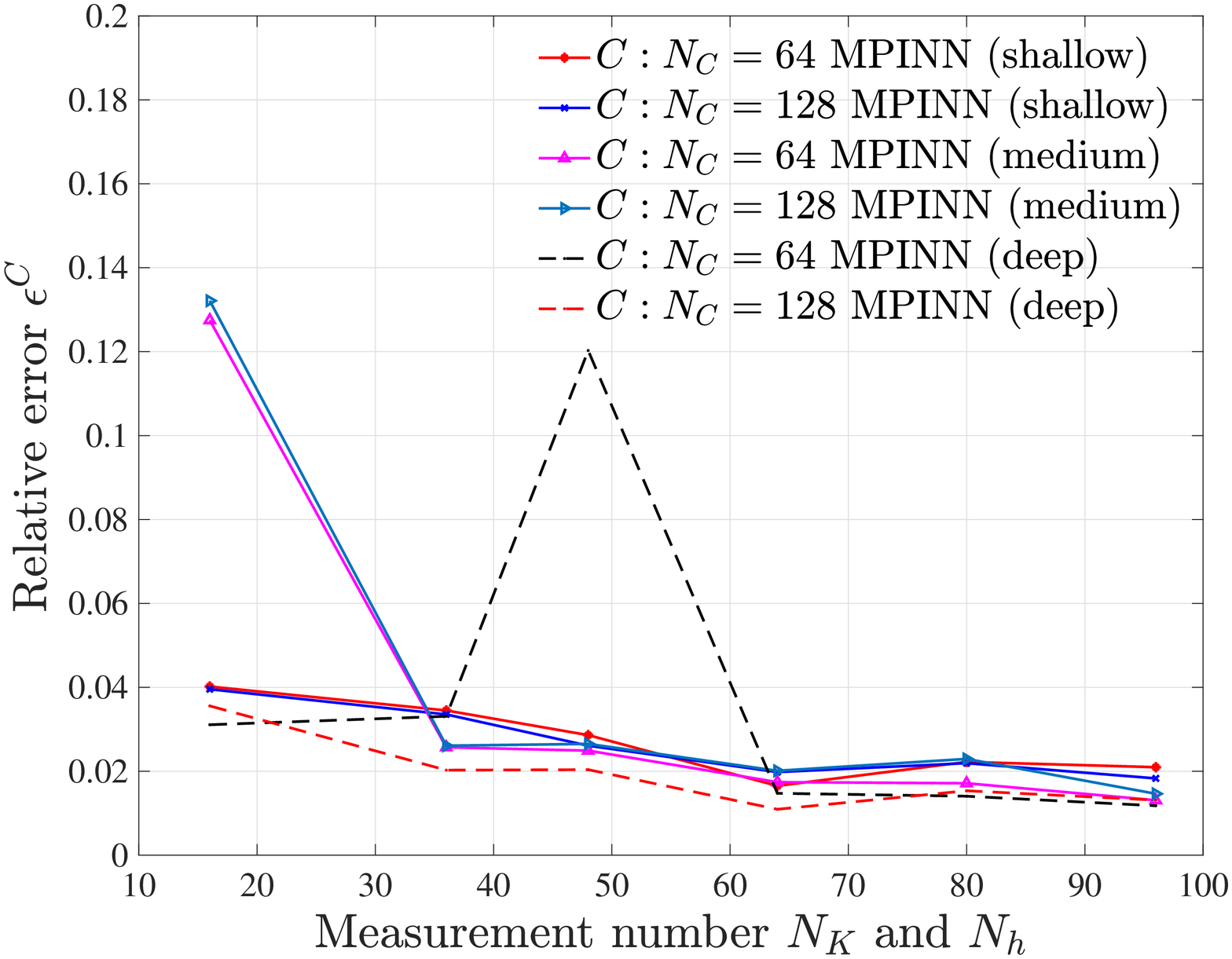}}
	\caption{The mean relative errors versus $N_K$ and $N_h$ in the estimated (a)  $K(\vec{x})$, (b)  $h(\vec{x})$, and (c) $C(\vec{x})$ using MPINN with $N_C = 0, 64, 128$ and the ``shallow'' [-32,32,32-], ``medium'' [-32,32,32,32-], and ``deep'' [-32,32,32,32,32-] DNNs $\hat{K}(\vec{x};\theta_K)$, $\hat{h}(\vec{x};\theta_h)$, and $\hat{C}(\vec{x};\theta_C)$. The PINN-Darcy estimation error for $K$ and $h$ are given for comparison.  The DNNs are trained with $N_f^h = 200$ and $N_f^C = 1000$ residual points. The simultaneous training is used in MPINNs.}
	\label{fig:MPINN-conv-st}
\end{figure}

It is also worth noting that adding physics constraints makes it more difficult and expensive to minimize the loss function. The results in Figure \ref{fig:MPINN-conv} are obtained using sequential training of DNNs in MPINN. The sequential training is designed to simplify minimization of the loss function with multiple PDE constraints. However, the sequential training also has the potential to introduce errors due to ``decoupling'' of the coupled physical processes. In Figure \ref{fig:MPINN-conv-st}, we show the mean relative error in the estimated $K$, $h$, and $C$ when the three DNNs are trained simultaneously.  We can see that $K$ and $h$ have a similar accuracy when trained simultaneously and sequentially. On the other hand, sequential training outperforms the simultaneous training for estimating the $C$ field. More specifically, $\epsilon^C$ is significantly larger in simultaneous training than in the sequential training for deep and medium-size DNNs with a small number of measurements. These results show that the sequential training helps minimize the loss function and does not introduce additional errors and instabilities.

%
%%%%%%%%%%%%%%%%%%%%%%%%%%%%%%%%%%%%%%%%%%%%%%%%%%%%
%%%     Discussion     %%%
%%%%%%%%%%%%%%%%%%%%%%%%%%%%%%%%%%%%%%%%%%%%%%%%%%%%
\section{Numerical example 2: lognormal conductivity fields with different correlation lengths}\label{sec:diss}
%
%
%%% Random field %%%
\subsection{Optimal DNN size as a function of the correlation length of the modeled field}\label{sec:correlation}

In the previous section, we demonstrated that the network size affects the accuracy of the DNN predictions, especially in the presence of sparse data. 
In this section, we investigate how to select the optimal DNN size for modeling spatially-correlated $K$ fields with different correlation lengths.

\begin{figure} [ht!]
	\centering
	\subfloat[] {\includegraphics[angle=0,width=2.2in]{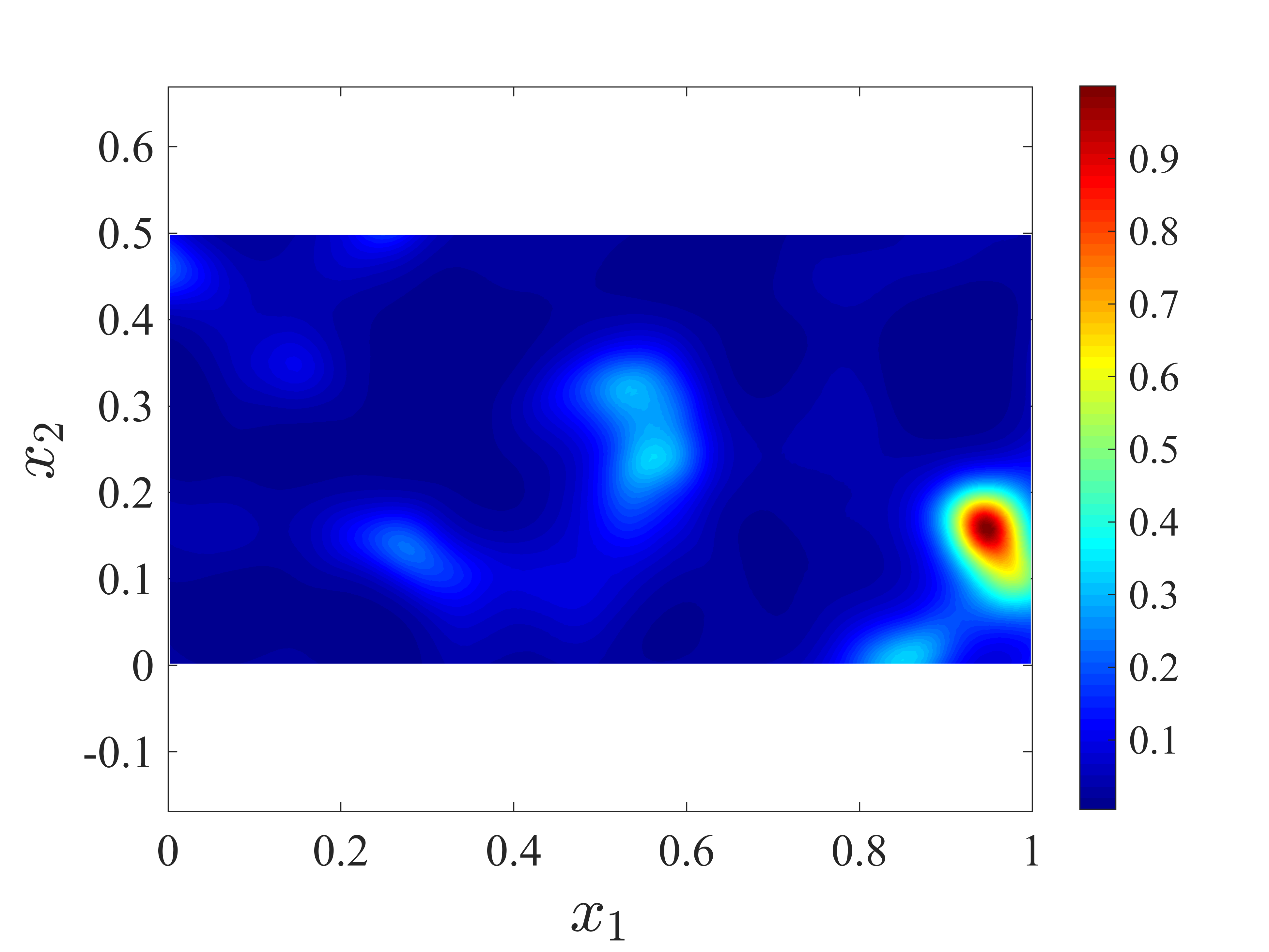}}
	\subfloat[] {\includegraphics[angle=0,width=2.2in]{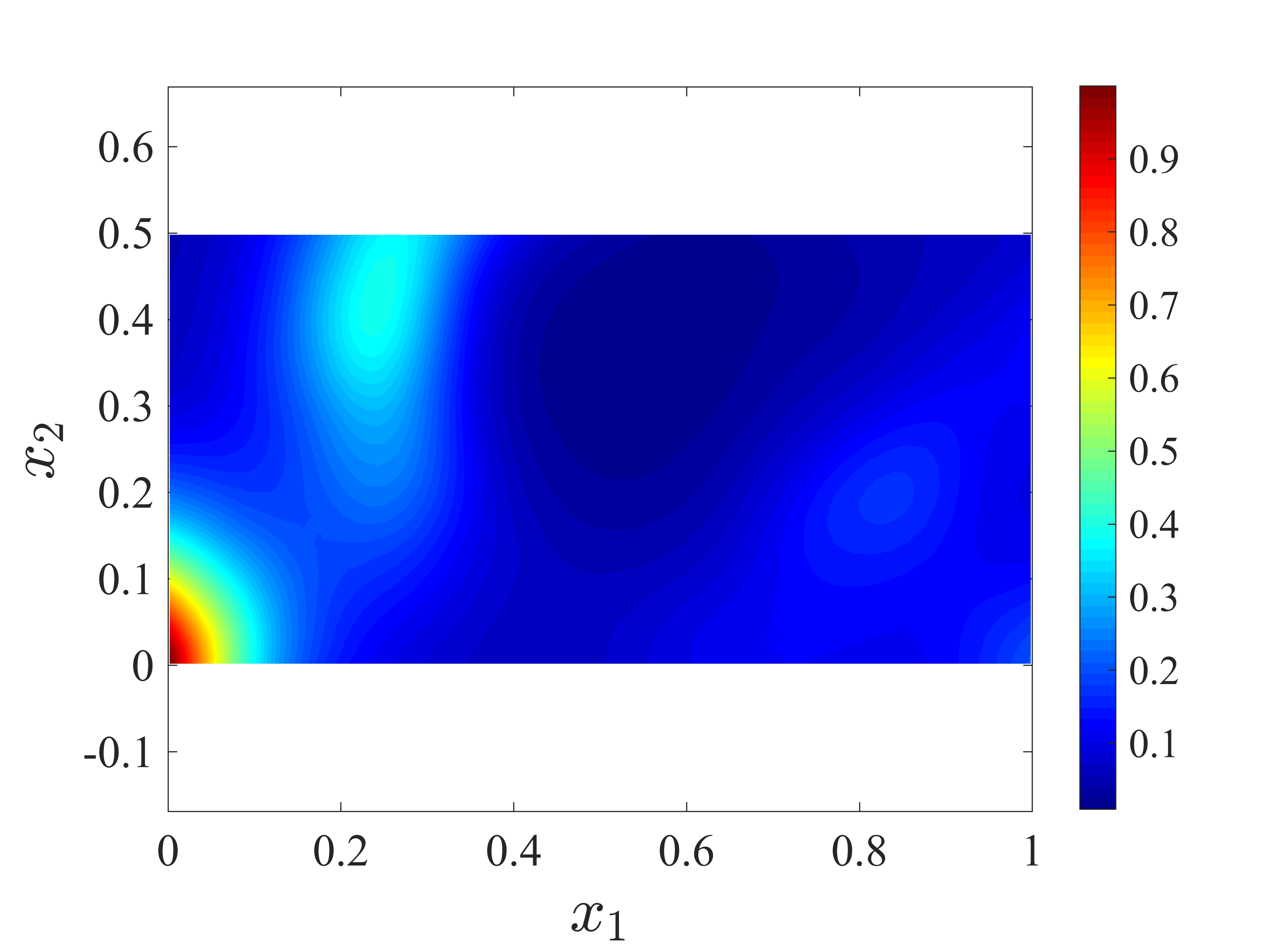}}
	
	\subfloat[] {\includegraphics[angle=0,width=2.2in]{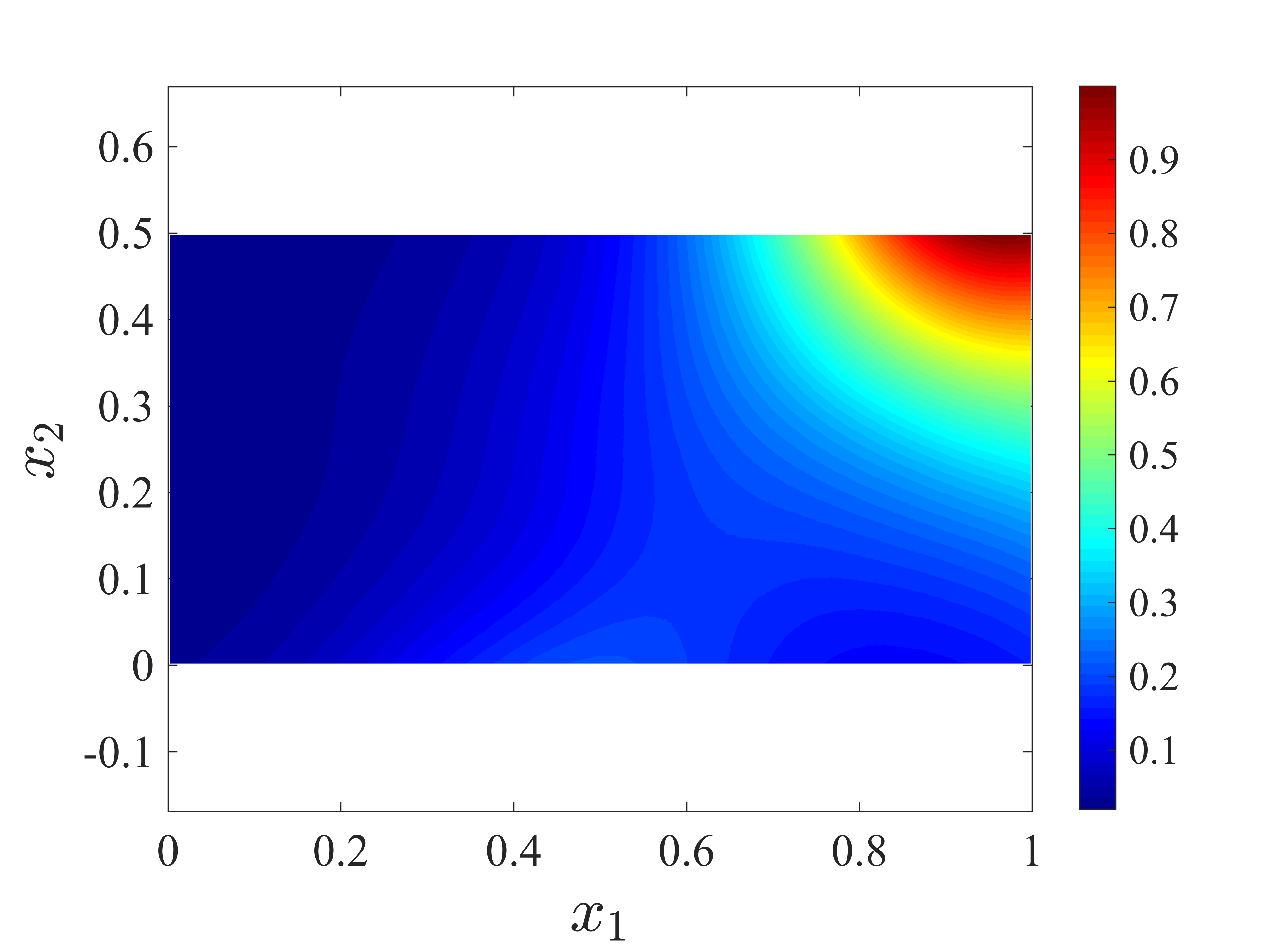}}
	\caption{Reference conductivity fields with the correlation length: (a)  $\lambda = 0.2$, (b)  $\lambda = 0.5$, and (c)  $\lambda = 1.0$.}
	\label{fig:conducitivity_corr}
\end{figure}

We examine three conductivity fields $K(\vec{x}) = \exp(Y(\vec{x}))$ obtained as realizations of the Gaussian field $Y(\vec{x})$ with the covariance function $C(\vec{x},\vec{x}')=\sigma^2 \exp(-||\vec{x}-\vec{x}'||/2\lambda^2)$ and the correlation lengths $\lambda = 0.2$, $\lambda = 0.5$, and $\lambda = 1.0$; see Figure \ref{fig:conducitivity_corr}.

\begin{table}[htb]
	\centering
	\caption{The number of total tunable parameters corresponding to the DNN architecture [2-$m_h$-$m_h$-$m_h$-1] as a function of $m_h$.}
	\begin{tabular}{ccccccccccccccc}
		\toprule
		$m_h$ & $10$    & $20$     & $30$      & $40$     & $50$ &  $60$ &  $70$ &  $80$ &  $90$ &  $100$\\
		\midrule
		DOF     & $261$  & $921$   & $1981$   & $3441$ &  $5301$  &  $7561$ &  $10221$ &  $13281$ &  $16741$ &  $20601$\\ 
		\bottomrule
	\end{tabular}
	\label{table:nn_size}
\end{table}

We first study the representative properties of DNNs as a function of the DNN size for these three conductivity fields.   
Here, we use the [2-$m_h$-$m_h$-$m_h$-1] $\hat{K}(\vec{x},\theta_K)$ DNN and vary $m_h$, which is the number of neurons in each hidden layer. The number of number of tunable parameters as a function of  $m_h$ for the chosen DNN architecture is given in Table \ref{table:nn_size}.
The conductivity fields in Figure \ref{fig:conducitivity_corr} are generated on the domain $\Omega=[0,1]\times[0,0.5]$ on a $256 \times 128$ grid with 32,768 grid points.  Here, we use values of $K$ at 20,000 grid points to train $\hat{K}(\vec{x},\theta_K)$ (without any physics constraints) and use $K$ values at all 32,768 grid points to evaluate the $\hat{K}(\vec{x},\theta_K)$ accuracy. 

For this  large number of measurements, we find that the L-BFGS-B algorithm is not efficient for minimizing the loss function, especially for the field with $\lambda = 0.2$.
Therefore, here we adopt the  Adam stochastic gradient descent (SGD) method~\cite{Kingma2014} with the experimentally determined initial learning rate of 0.0002 and the batch size of 1000. 
We find that $4 \times 10^5$ iterations are needed to train $\hat{K}(\vec{x},\theta_K)$ for $\lambda = 0.2$, $3 \times 10^5$ for $\lambda = 0.5$, and $2 \times 10^5$ for $\lambda = 1.0$ to achieve a sufficiently low training error.
Our results show that less iterations are needed to train DNNs for smoother conductivity fields with larger correlation lengths.

\begin{figure} [ht!]
	\centering
	\subfloat[Correlation length $0.2$] {\includegraphics[angle=0,width=2.2in]{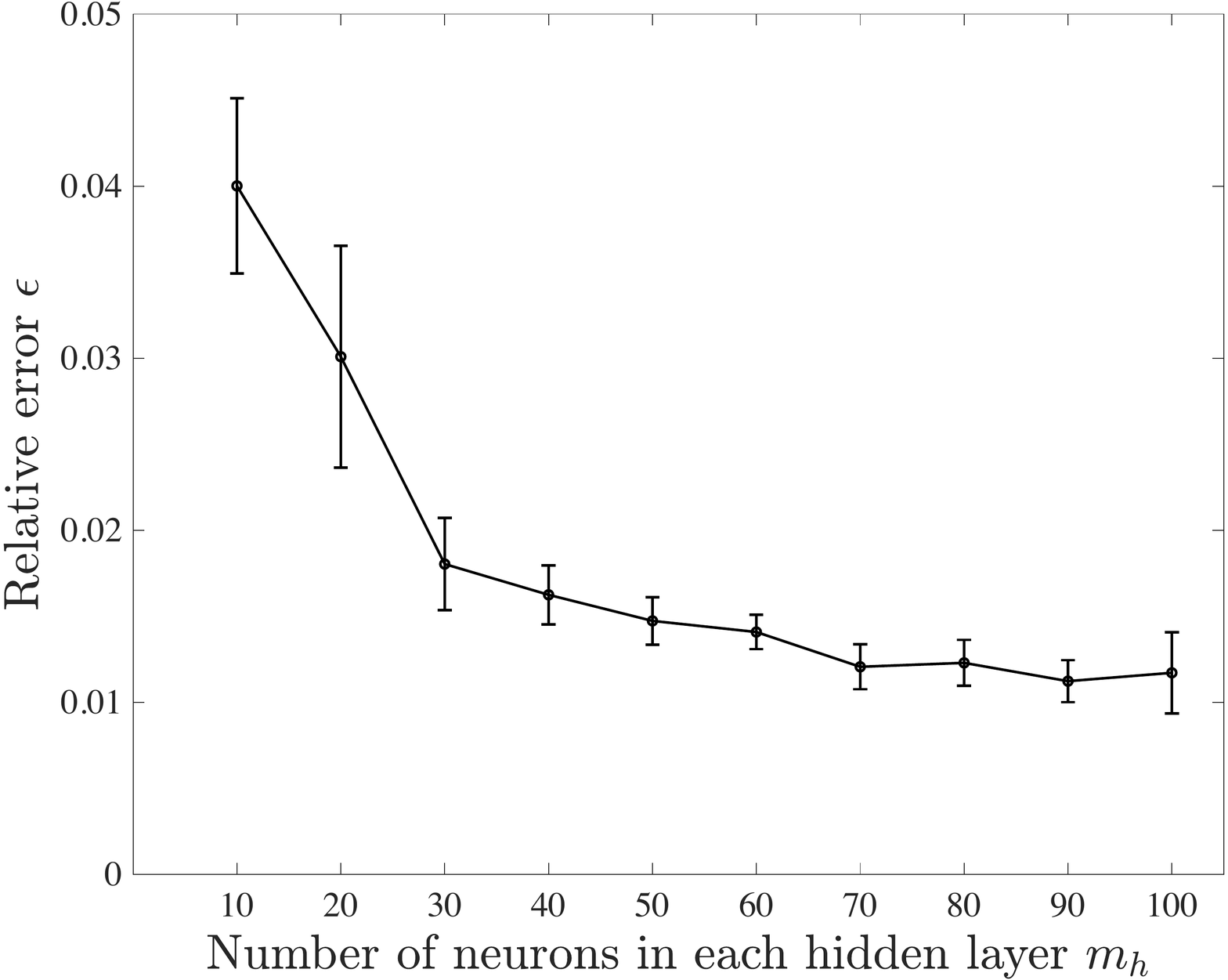}}
	\subfloat[Correlation length $0.5$] {\includegraphics[angle=0,width=2.2in]{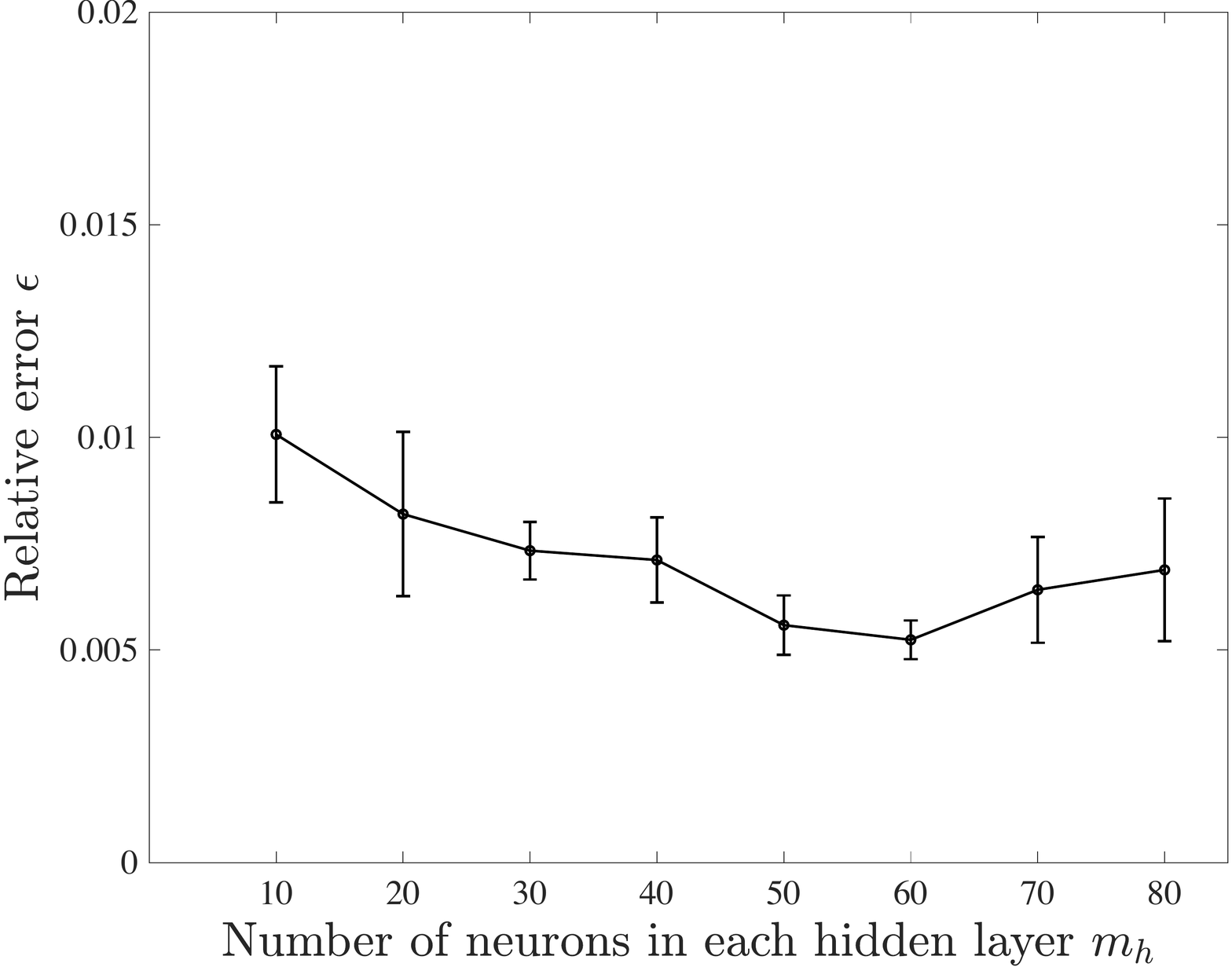}}
	
	\subfloat[Correlation length $1.0$] {\includegraphics[angle=0,width=2.2in]{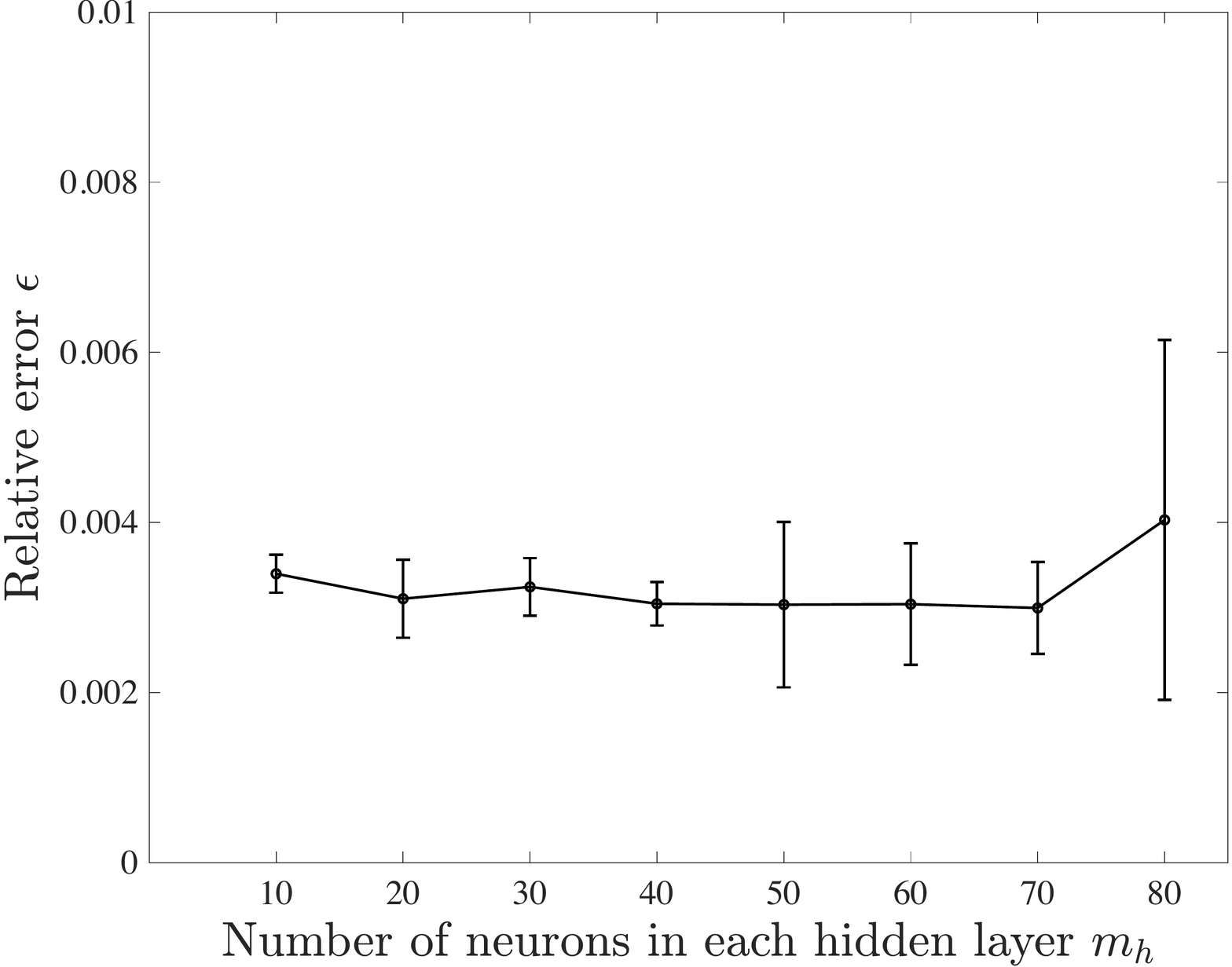}}
	\caption{The relative errors of DNN approximation against the increase of network size for three conductivity fields of the correlation length: (a) $\lambda = 0.2$, (b) $\lambda = 0.5$, and (c) $\lambda = 1.0$. The abscissa corresponds to the number of neurons in each hidden layers.}
	\label{fig:res_err_vs_nn}
\end{figure}

%\subsubsection{DNN approximation results}
Figure \ref{fig:res_err_vs_nn} shows the mean relative error of $\hat{K}(\vec{x},\theta_K)$ and the standard deviation as a function of $m_h$ for the three correlation lengths. The mean errors and standard deviation of the error are computed  from simulations with 10 different DNN initializations. 
Initially, the approximation error decreases as the DNN size increases because of the increasing DNN representation ability. We can see that a smaller DNN is sufficient to represent a smoother field with larger correlation length. We also see that for fields with the correlation length 0.5 and 1, the approximation error increases due to overfitting once the DNN size exceeds the ``optimal size.''  
For example, for the $K$ field with $\lambda = 0.5$ (see Figure \ref{fig:res_err_vs_nn} (b)), the smallest relative error of $0.52 \%$ is reached at $m_h=60$. 
DNNs with $m_h<60$  are not representative enough, and DNNs with $m_h>60$ cause overfitting. Therefore, we postulate that the DNN with three hidden layers and $m_h=60$ is the optimal-size DNN for this $K$ field.   For the $K$ fields with  $\lambda = 0.2$ and $\lambda = 1.0$, the optimal DNN size is reached at $m_h=90$ and $m_h=40$, respectively. 
For the $K$ field with $\lambda = 1.0$, the minimum of the mean error function ($\approx 0.3\%$) is very shallow, as shown in Figure \ref{fig:res_err_vs_nn} (c). Hence, we select $m_h=40$ as the optimal DNN width because it results in the smallest error standard deviation. 

\begin{figure}[ht!]
	\centering
	\includegraphics[angle=0,width=3in]{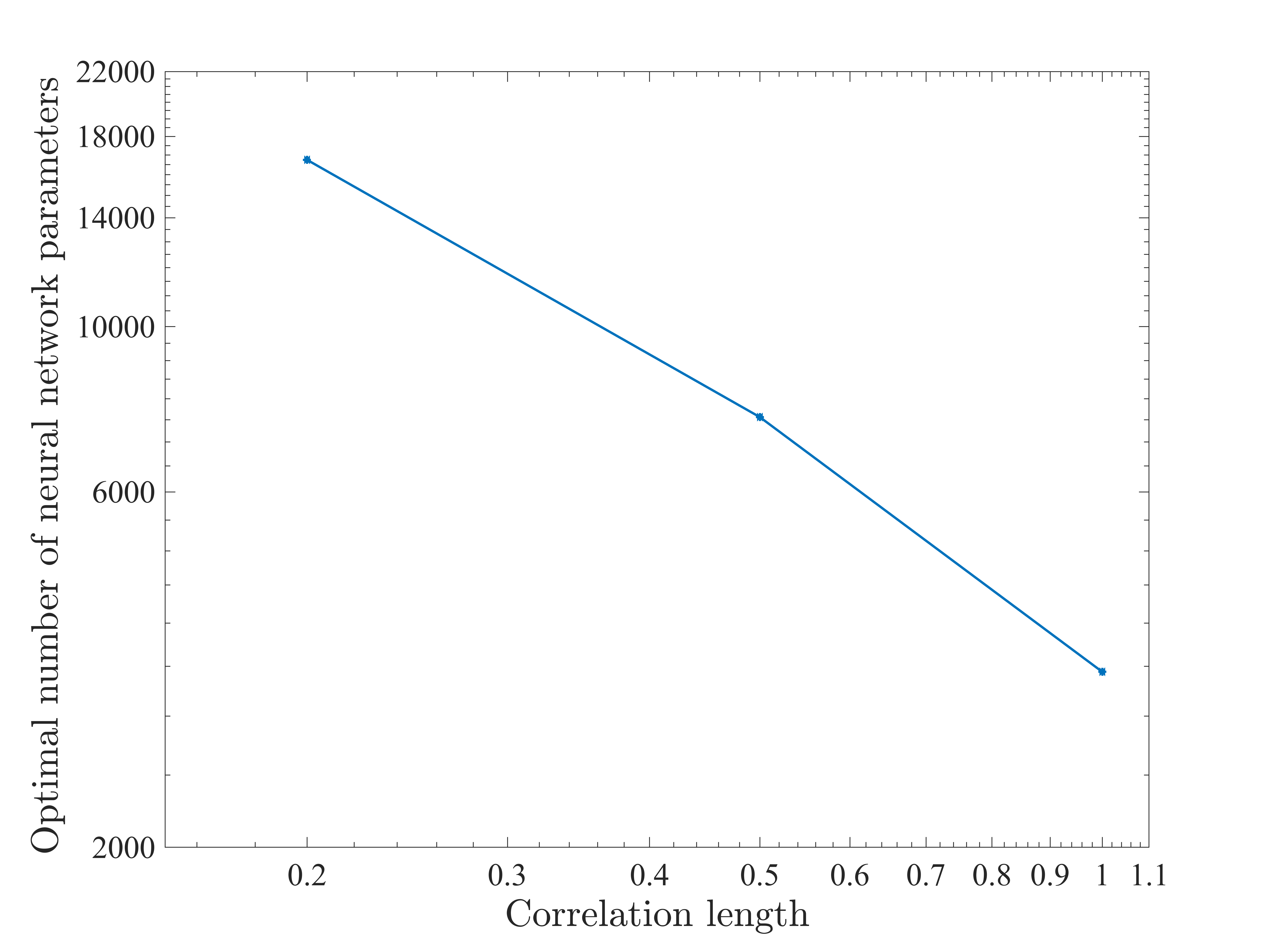}
	\caption{The logarithmic plot of the optimal neural network size against the correlation length used for generating conductivity fields.}
	\label{fig:res_nn_vs_corr}
\end{figure}

Figure \ref{fig:res_nn_vs_corr} shows an approximately power-law dependence of the optimal DNN size (in terms of the total number of DNN parameters) as a function of the correlation length of the modeled field. 
It is important to note that  in addition to the correlation length of the modeled field, the optimal DNN size  depends on many other factors, including the type of activation function and the number of hidden layers. In this study, we fix the number of hidden layers and the activation function. Therefore, the results in Figure  \ref{fig:res_nn_vs_corr}  might not apply to other DNN architectures. 

%
%
%%% Random field %%%
\subsection{Data assimilation}\label{data_assimilation}
\begin{figure} [ht!]
	\centering
	\subfloat[Correlation length $0.2$] {\includegraphics[angle=0,width=2.2in]{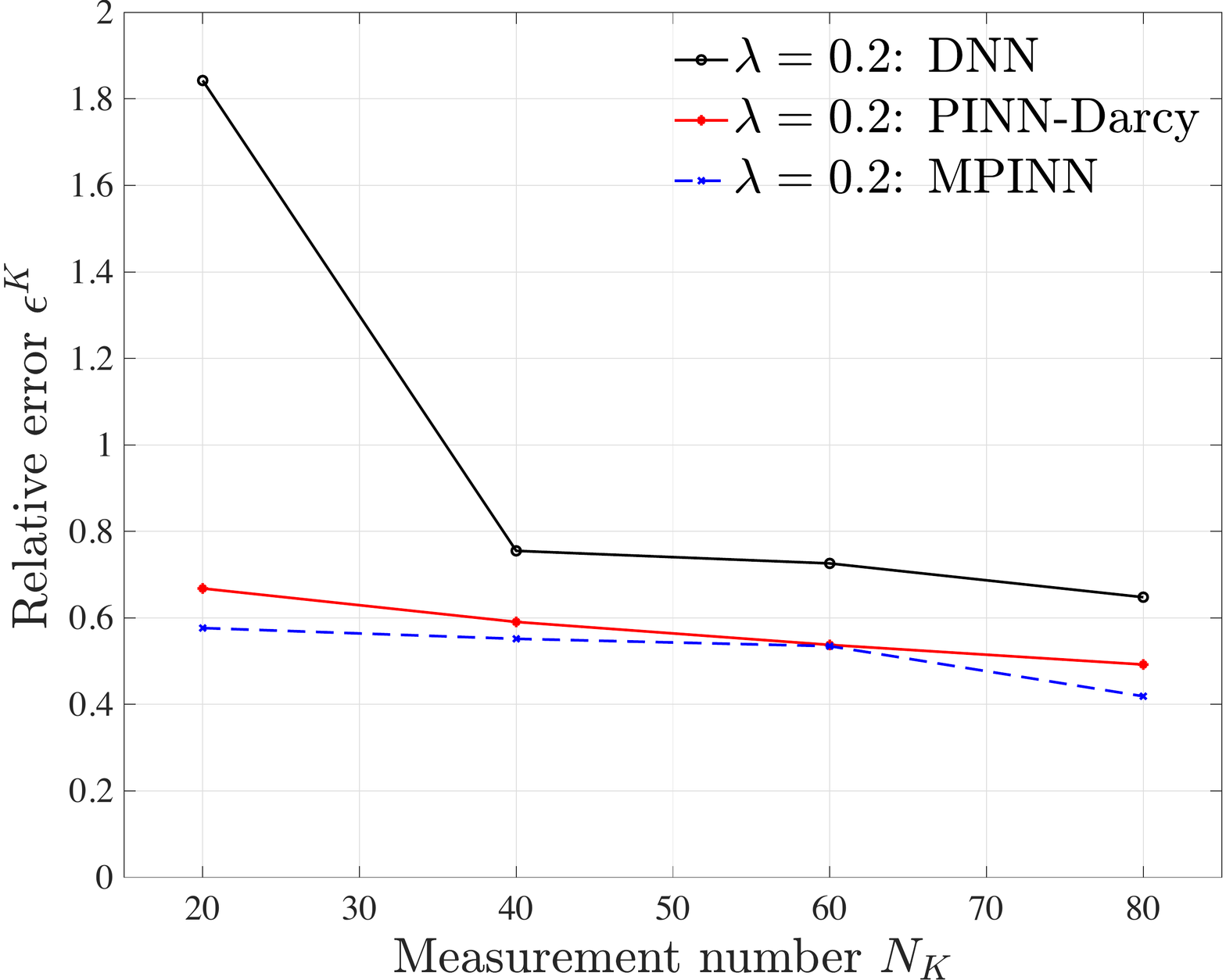}}
	\subfloat[Correlation length $0.5$] {\includegraphics[angle=0,width=2.2in]{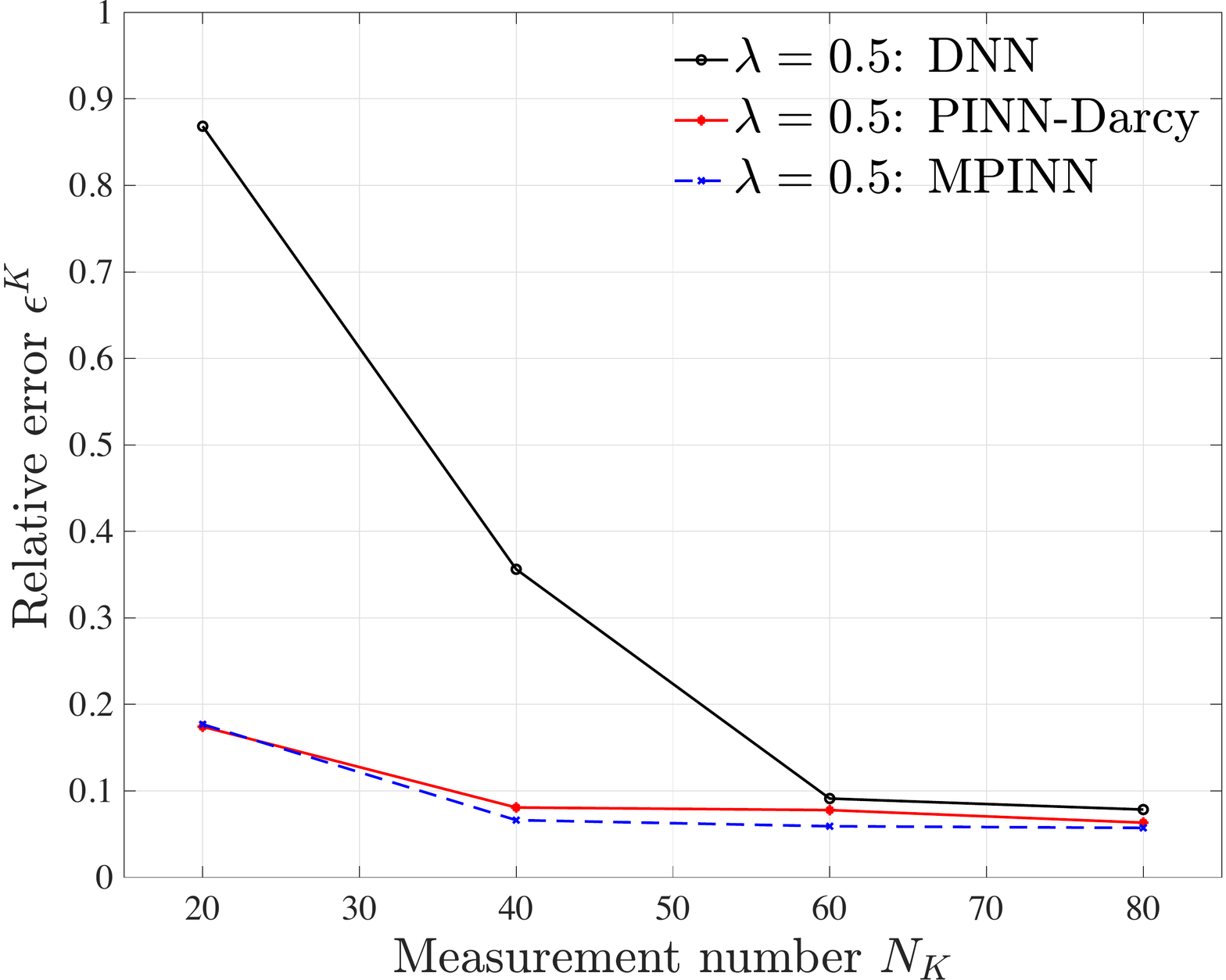}}
	
	\subfloat[Correlation length $1.0$] {\includegraphics[angle=0,width=2.2in]{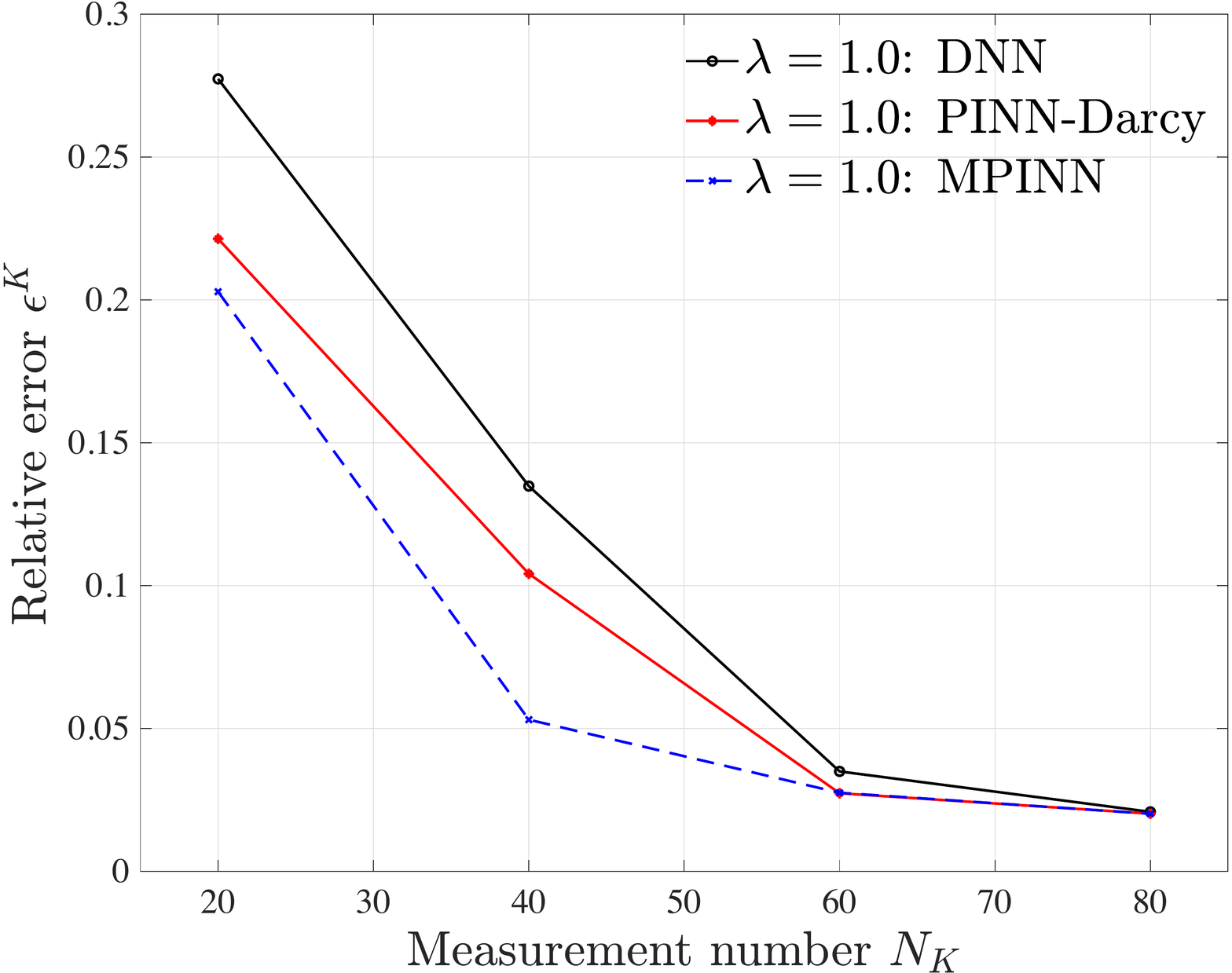}}
	\caption{The relative errors against the increase of measurements $N_K$ (with fixing $N_h = 40, N_C = 100, N_f^h = 1000, N_f^C = 1000$) for approximating the conductivity $K(\vec{x})$ fields generated by different correlation length: (a) $\lambda = 0.2$, (b) $\lambda = 0.5$, and (c) $\lambda = 1.0$. The approximations of the DNN, PINN-Darcy, and MPINN approaches are obtained by using the network structure [2-$m_h$-$m_h$-$m_h$-1] and $m_h=60$.}
	\label{fig:rand_conv_k}
\end{figure}

Next, we compare the data-driven DNN, PINN-Darcy, and MPINN methods for data assimilation and parameter and state estimation using $K$, $h$, and $C$ measurements.  The $K$ fields shown in Figure \ref{fig:conducitivity_corr} are used as the ground truth. The ground truth $h$ and $C$ fields are generated as the solutions of the Darcy and advection--dispersion equations using the STOMP code. Based on the analysis in Section \ref{sec:correlation}, we first use the DNN structure of [2-$m_h$-$m_h$-$m_h$-1] with $m_h = 60$ for $\hat{K}$, as this network size produces a reasonable  (but not optimal) fit for these three correlation lengths. To optimize the loss function, we adopt the Adam algorithm with the learning rate of 0.0002, then switch to the L-BFGS method until the training loss reaches a low threshold of 0.0005. 

Figure \ref{fig:rand_conv_k} compares the approximation errors in the data-driven DNN, PINN-Darcy, and MPINN methods as a function of the number of $K$ measurements for the three conductivity fields.  In these simulations, we use $N_h = 40$ and $N_C = 100$ measurements of head and concentration, respectively. The number of residual points in the PINN-Darcy method is $N_f^h = 1000$ and $N_f^C = 0$, and  $N_f^h = 1000$ and $N_f^C = 1000$ in the MPINN method. The number of residual points in the data-driven method is zero. 

\begin{table}[htb]
	\centering
		\caption{Relative errors in the estimated $K$,  $h$, and $C$ versus $N_K$ for $\lambda = 0.2, 0.5$, and 1.0. Here, $N_h = 40$, $N_C = 100$, $N_f^h = 1000$, and $N_f^C = 1000$. The DNN architecture is [2-$m_h$-$m_h$-$m_h$-1] with $m_h=60$ for all $\hat{K}$, $\hat{h}$, and $\hat{C}$.}
	\begin{tabular}{c|ccccc|cccc}
		\toprule
		\multicolumn{2}{c|}{} & \multicolumn{4}{|c|}{$\epsilon^h$} &  \multicolumn{4}{|c}{$\epsilon^C$} \\
		\hline
		\multicolumn{2}{c|}{$N_K$} 	& 20    & 40   & 60    & 80    & 20    & 40   & 60    & 80 \\
		\hline
		\multirow{3}{*}{$\lambda=0.2$} 
		& \multicolumn{1}{c|}{DNN}        & \multicolumn{4}{c|}{$4.72 \%$} & \multicolumn{4}{c}{$18.25 \%$} \\
		& \multicolumn{1}{c|}{PINN} 	  & $6.82 \%$    & $6.49 \%$   & $4.74 \%$    & $4.35 \%$    &    &    &     &  \\
		& \multicolumn{1}{c|}{MPINN} 	& $6.71 \%$    & $6.39 \%$   & $5.19 \%$    & $3.74 \%$  & $8.60 \%$    & $7.35 \%$   & $5.91 \%$    & $7.10 \%$ \\
		\hline
		\multirow{3}{*}{$\lambda=0.5$} 
		& \multicolumn{1}{c|}{DNN} 	 & \multicolumn{4}{c|}{$1.75 \%$} & \multicolumn{4}{c}{$18.65 \%$} \\
		& \multicolumn{1}{c|}{PINN} 	  & $0.94 \%$    & $0.92 \%$   & $0.75 \%$    & $0.57 \%$    &    &    &     &  \\
		& \multicolumn{1}{c|}{MPINN} 	& $1.04 \%$    & $0.69 \%$   & $0.75 \%$    & $0.48 \%$    & $2.02 \%$    & $1.14 \%$   & $1.41 \%$    & $1.36 \%$ \\
		\hline
		\multirow{3}{*}{$\lambda=1.0$} 
		& \multicolumn{1}{c|}{DNN} 	 & \multicolumn{4}{c|}{$1.28\%$} & \multicolumn{4}{c}{$16.72 \%$} \\
		& \multicolumn{1}{c|}{PINN} 	  & $6.43 \%$    & $2.58 \%$   & $0.72 \%$    & $0.60 \%$    &    &    &     &  \\
		& \multicolumn{1}{c|}{MPINN} 	& $2.53 \%$    & $0.95 \%$   & $0.74 \%$    & $0.63 \%$    & $3.51 \%$    & $1.20 \%$   & $1.13 \%$    & $1.35 \%$ \\
		\bottomrule
	\end{tabular}
	\label{table:rand_conv}
\end{table}

For all three correlation lengths, the approximation error decreases with increasing $N_K$. We also see that adding physics constraints improves the accuracy of the DNN approximation of the $K$ field. The biggest reduction in the estimation error in archived by adding $h$ measurements and the Darcy equation constraint, as evident from the comparison of the data-driven DNN and PINN-Darcy estimation errors. Adding $C$ measurements and the advection--dispersion equation constraints further reduces the approximation error. The reduction of errors due to assimilating different data and physics is especially pronounced for sparse data (small $N_K$) and small correlation lengths.  For $\lambda = 0.2$ and $N_K=20$ (see Figure \ref{fig:rand_conv_k} (a)), the relative error of parameter estimation drops from more than $180 \%$ when using the data-driven DNN method to $66.8 \%$ in the PINN-Darcy estimate to $57.7 \%$ in the MPINN estimate.

MPINN also provides a significantly improved hydraulic head and concentration estimations as compared to the data-driven DNN method, as evident in the comparison of the mean errors $\epsilon^h$ and $\epsilon^C$ in Table \ref{table:rand_conv}. 
We note that the number of measurements $N_h$ and $N_C$ is fixed, and the data-driven DNN hydraulic head and concentration estimates do not depend on $N_K$.  
On the other hand,  the PINN-Darcy and MPINN estimations of $h$ and $C$ improve with increasing $N_K$. This demonstrates the capability of the physics-informed DNNs to learn from indirect measurements. 

The improvements due to using physics constraints and indirect measurements is particularly pronounced for estimating (highly nonlinear) $C(\vec{x})$ can be seen in the comparison of the data-driven and MPINN $\epsilon^C$ estimation errors.  For example, for $\lambda=1$ and $N^C=100$, $\epsilon^C$ decreases from 16.72\% in the data-driven DNN to 1.35\%. 

As we previously observed in Section \ref{sec:PINN_Darcy}, the effect of enforcing physics for the much smoother and nearly linear $h$ field is less pronounced. Here, with 40 $h$ measurements, the data-driven DNN can be trained relatively accurately, and adding physics constraints slightly increases the mean $\epsilon^h$ error unless a sufficiently large number of  $N_K$ measurements are available. Possible reasons for a slight increase in errors due to adding physics constraints when a large number of direct measurements is available are listed in Section \ref{sec:PINN_Darcy}.

\begin{figure} [ht!]
	\centering
	\subfloat[$\epsilon^K=35.62\%$] {\includegraphics[angle=0,width=2.2in]{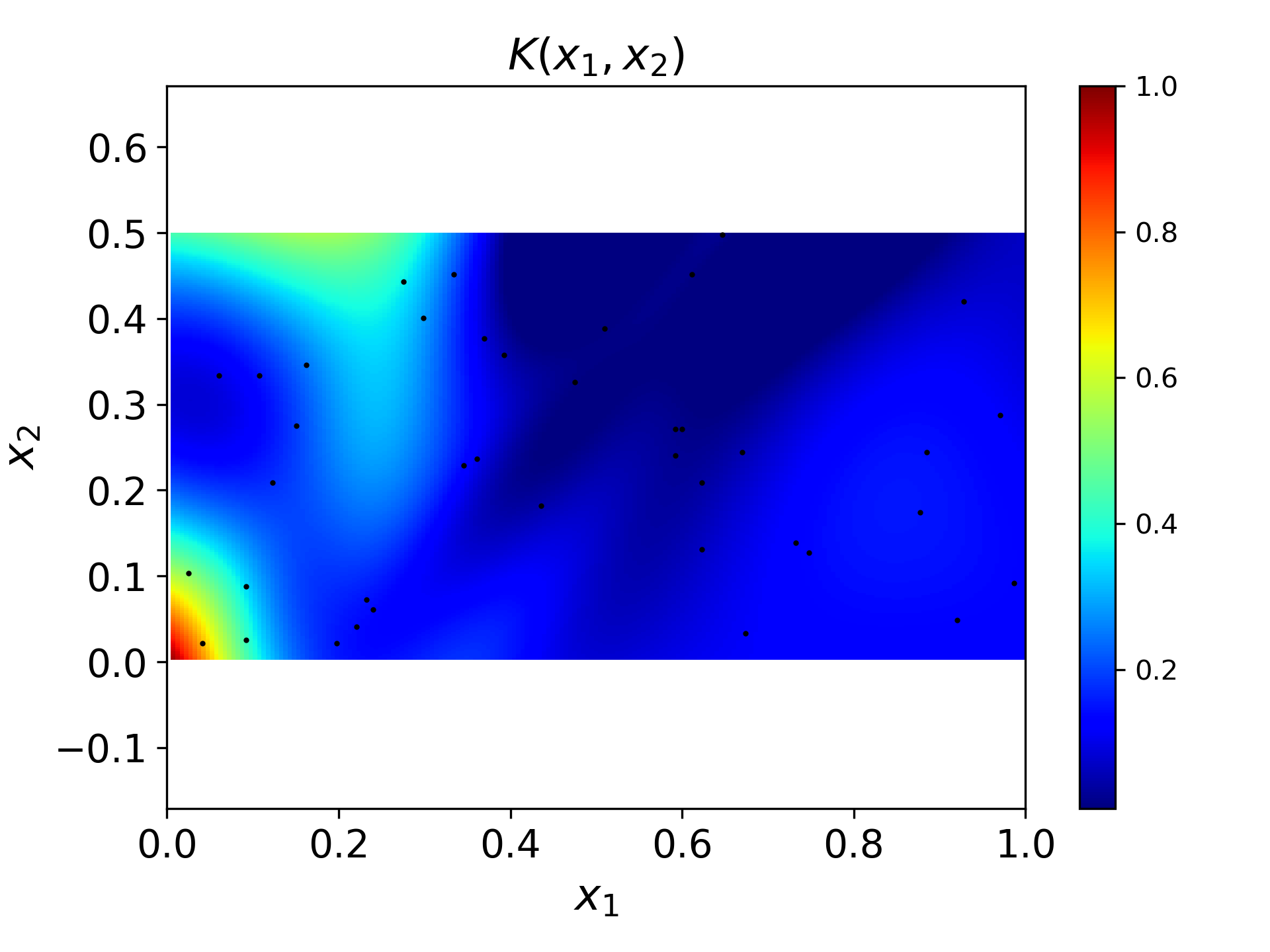}}
	\subfloat[$\epsilon^K=8.08\%$] {\includegraphics[angle=0,width=2.2in]{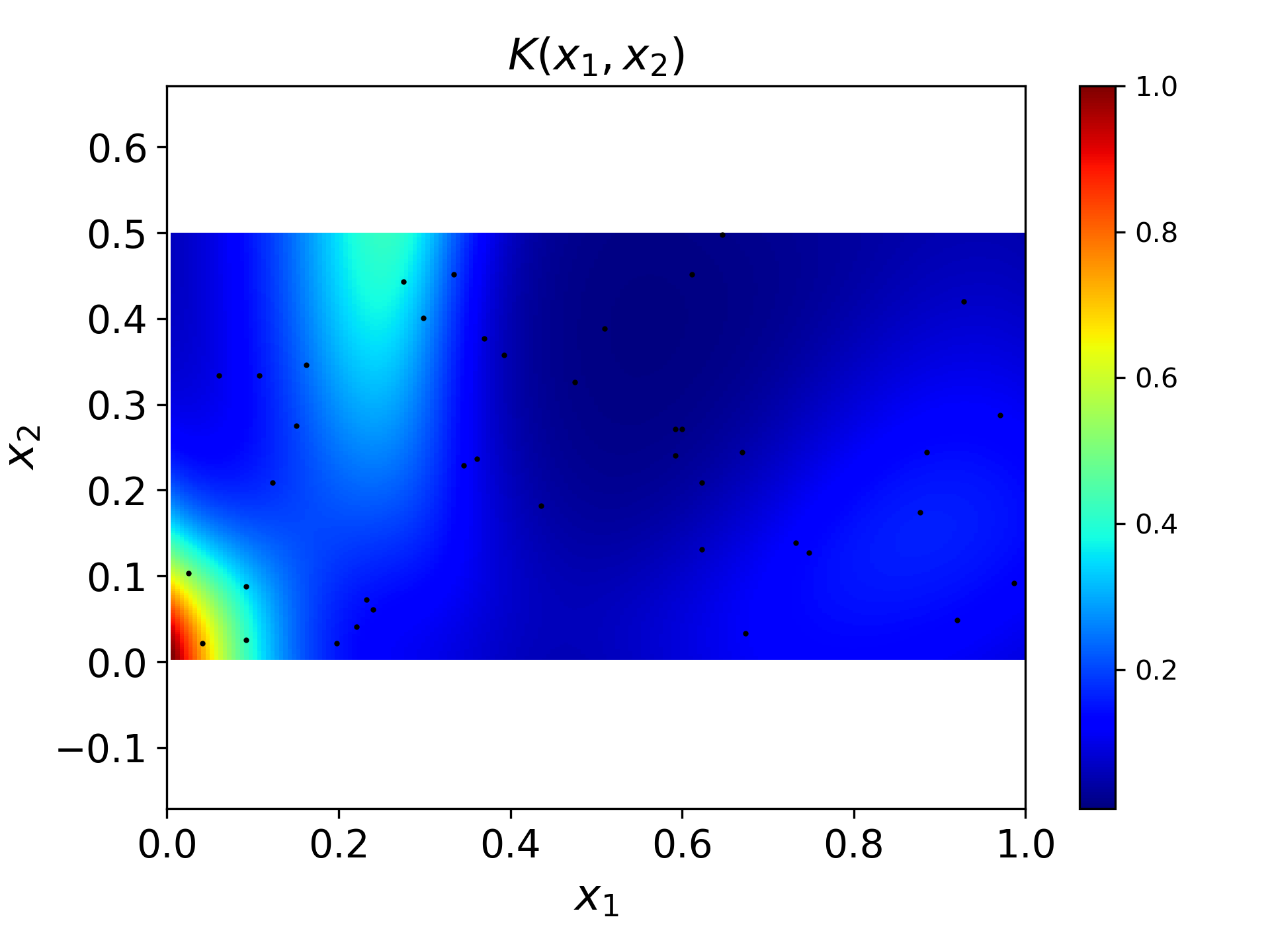}}
	
	\subfloat[$\epsilon^K=6.62\%$] {\includegraphics[angle=0,width=2.2in]{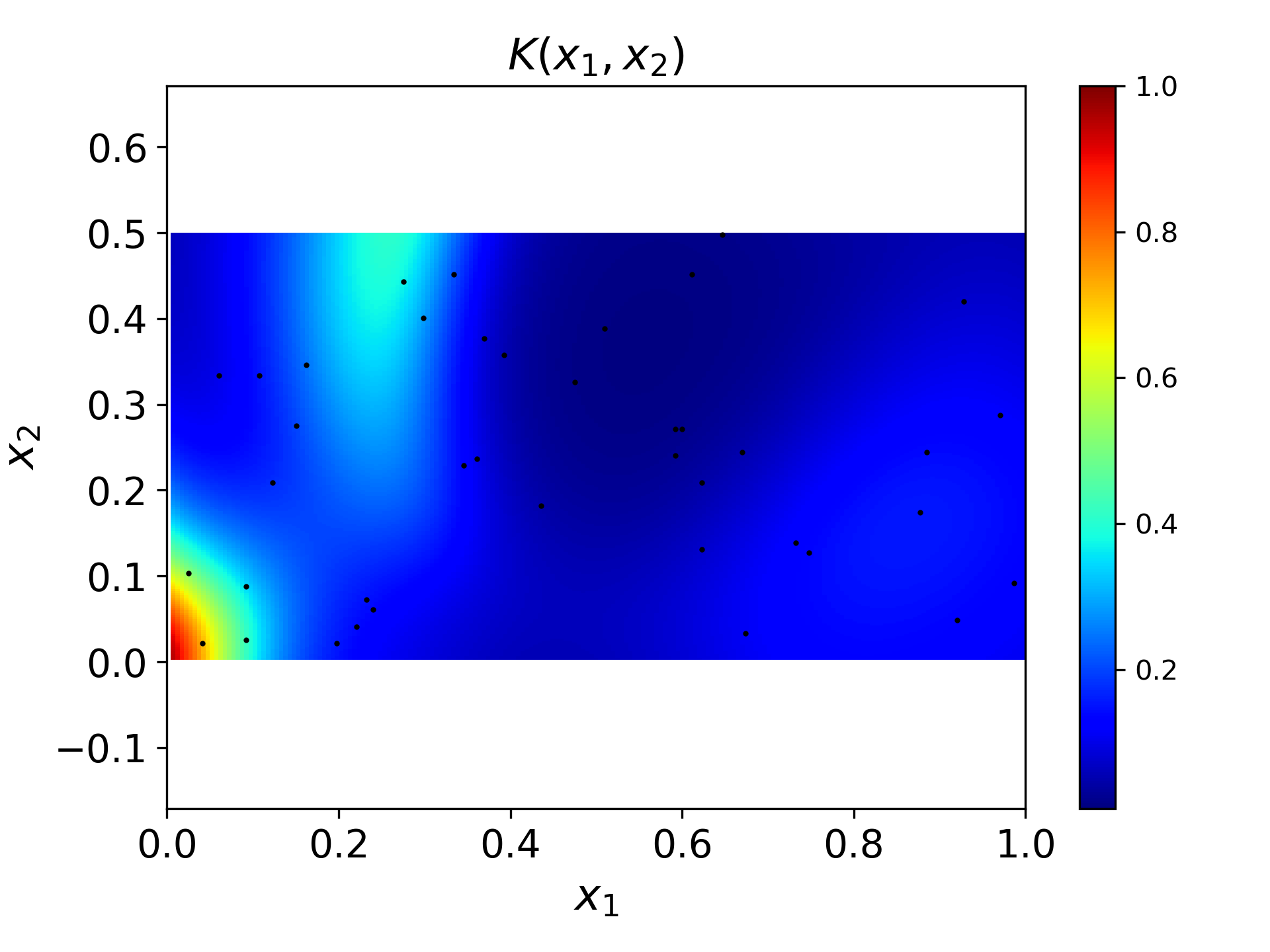}}
	\caption{$K$ field ($\lambda = 0.5$) estimated with (a) data-driven DNN, (b) PINN-Darcy, and (c) MPINN. The locations of $K$ measurements are denoted by black dots.}
	\label{fig:rand_2D_k}
\end{figure}

\begin{figure} [ht!]
	\centering
	\subfloat[Reference] {\includegraphics[angle=0,width=2.2in]{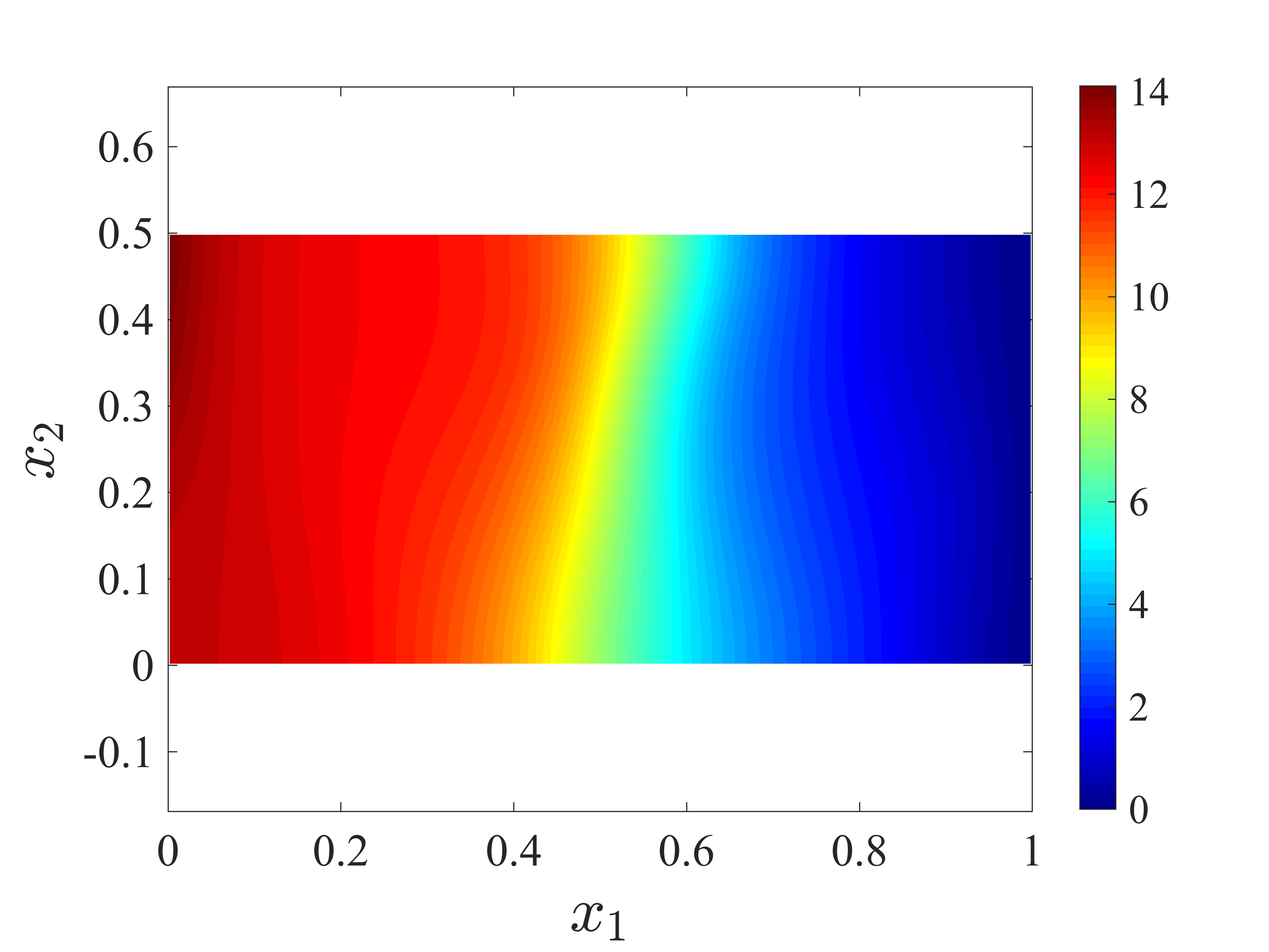}}
	\subfloat[$\epsilon^h=1.75\%$] {\includegraphics[angle=0,width=2.2in]{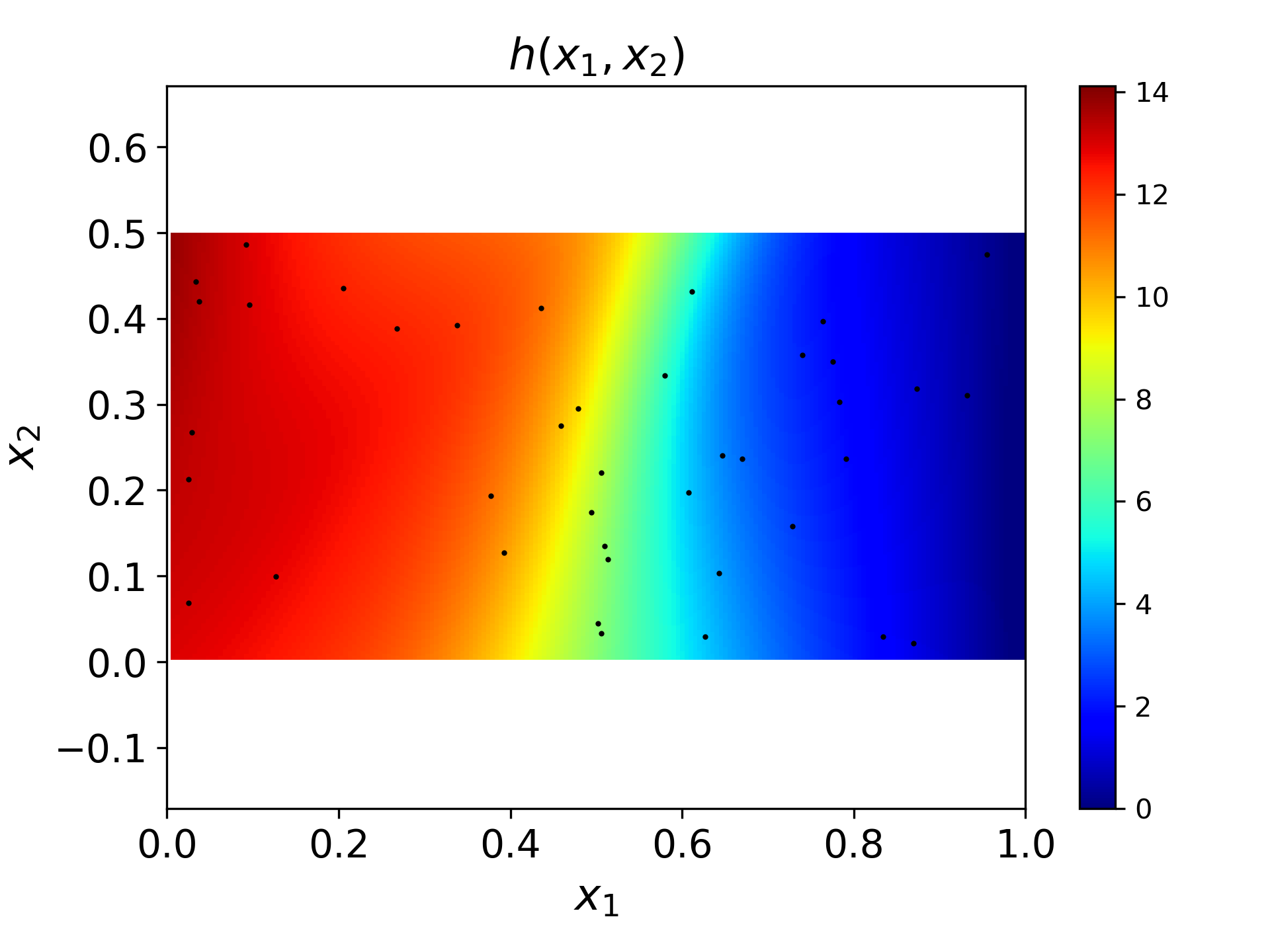}}\\
	\subfloat[$\epsilon^h=0.92\%$] {\includegraphics[angle=0,width=2.2in]{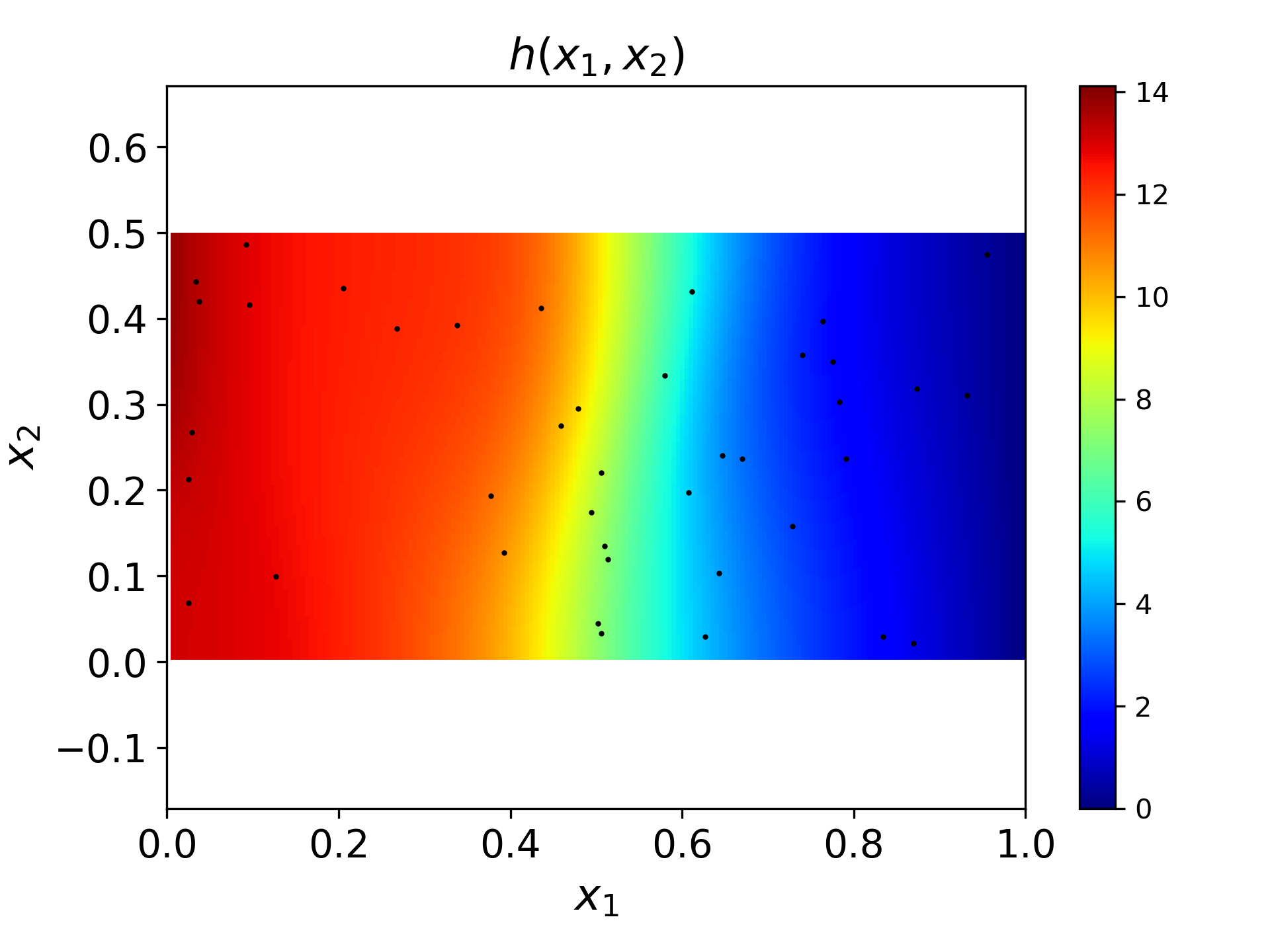}}
	\subfloat[$\epsilon^h=0.69\%$] {\includegraphics[angle=0,width=2.2in]{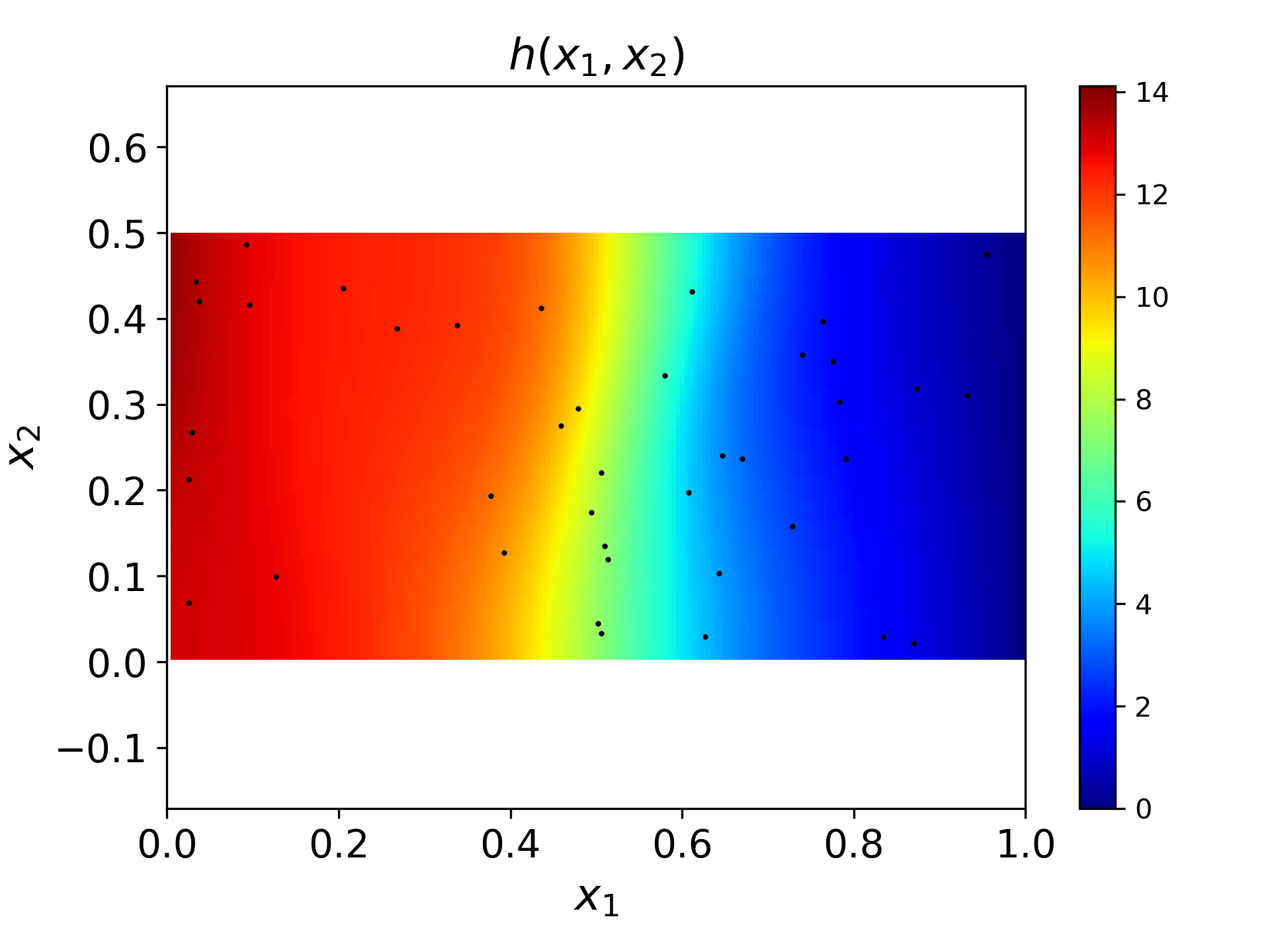}}
	\caption{ Hydraulic head $h$ field: (a) reference, (b) estimated with data-driven DNN, (c) estimated with PINN-Darcy, and (d) estimated with MPINN. The locations of $h$ measurements are denoted by black dots.}
	\label{fig:rand_2D_h}
\end{figure}

\begin{figure} [ht!]
	\centering
	\subfloat[Reference] {\includegraphics[angle=0,width=2.2in]{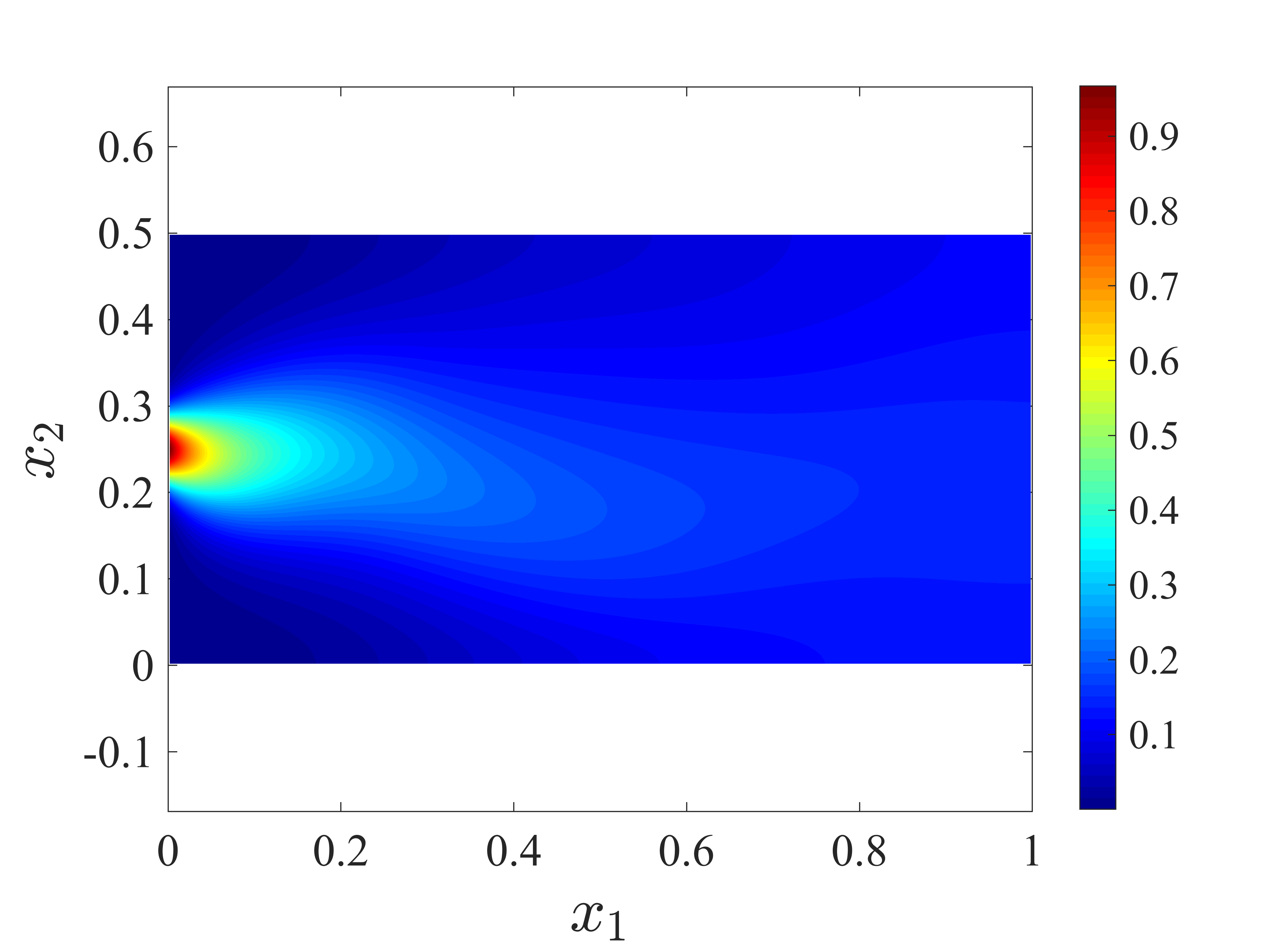}}
	\subfloat[$\epsilon^C=18.65\%$] {\includegraphics[angle=0,width=2.2in]{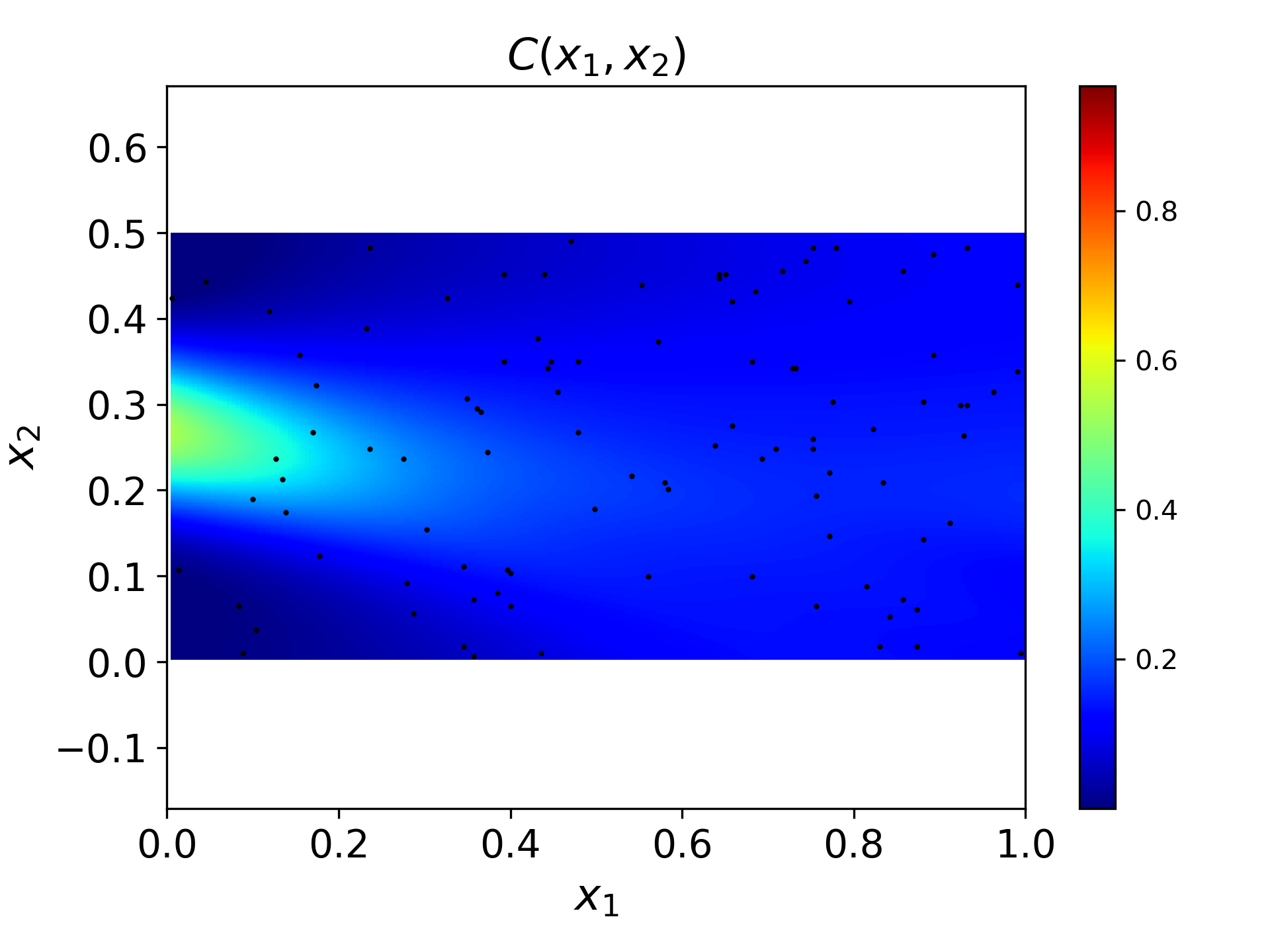}}
	
	\subfloat[$\epsilon^C=1.14\%$] {\includegraphics[angle=0,width=2.2in]{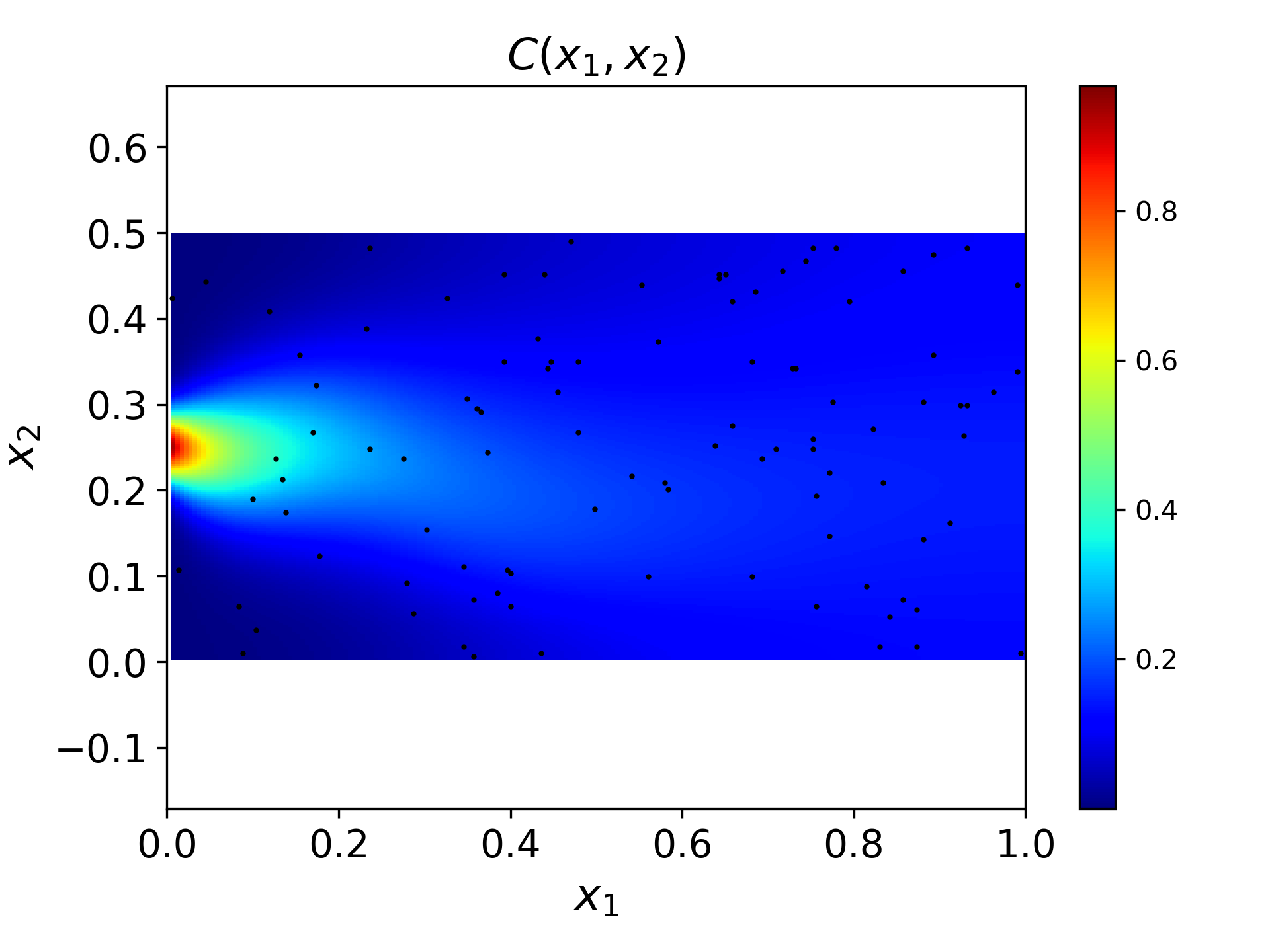}}
	\caption{Concentration field: (a) reference, (b) estimated with the data-driven DNN, and (c) estimated with MPINN. The locations of measurements are denoted by black dots.}
	\label{fig:rand_2D_c}
\end{figure}

For $\lambda = 0.5$,  in figures \ref{fig:rand_2D_k}--\ref{fig:rand_2D_c} we show the $K(\vec{x})$, ${h}(\vec{x})$, and ${C}(\vec{x})$ fields estimated with the data-driven DNN, PINN-Darcy, and MPINN methods, where $N_K = 40$, $N_h = 40$, $N_C = 100$, $N_f^h = 1000$, and $N_f^C = 1000$.  The comparison of the estimated $K$ fields in Figure \ref{fig:rand_2D_k} with the reference $K$ field in Figure \ref{fig:conducitivity_corr} (b) shows that PINN-Darcy significantly improves the data-driven DNN prediction. The MPINN further improves the $K$ estimation as indicated by the smaller $\epsilon^K$.  The data-driven DNN approximation near the upper left corner significantly differs from the ground truth $K$ field due to the lack of measurements in this region. However, the approximation error around this area is greatly reduced in the PINN-Darcy and MPINN methods, which leverages indirect observations (i.e., head and concentration observations) located in this area, as shown in figures \ref{fig:rand_2D_k} (b) and (c).

Although a good approximation of $h$ can be obtained with all three methods (see Figure \ref{fig:rand_2D_h}), we still observe that PINN and MPINN improve the data-driven DNN $h$ estimation as indicated by the smaller mean $\epsilon^h$ errors. For the highly-nonlinear $C$ field, the  data-driven DNN estimate is significantly less accurate than MPINN, both in terms of the concentration distribution and the mean $\epsilon^C$  error, as seen in Figure \ref{fig:rand_2D_c}.  Notably,  MPINN is able to accurately  describe the eye of the concentration plume with very few direct  measurements near this region. Once again, this demonstrates that  MPINN can use sparse direct and indirect measurements in combination with PDEs to capture local features that otherwise cannot be described with only direct (sparse) measurements.  

\subsubsection{The effect of DNN size on the estimation error}

Finally, we investigate whether using the optimal size $\hat{K}$, as determined in Section \ref{sec:correlation}, would reduce the error in the estimated $K$, $h$, and $C$ fields. As an example, we choose the case  with $\lambda=0.2$. According to Figure \ref{fig:res_err_vs_nn}(a), the optimal $\hat{K}$ size to represent a $K$ field with this correlation length is $m_h=90$. In Section  \ref{data_assimilation}, we used $\hat{K}$ with $m_h=60$ to estimate $K$, $h$, and $C$. Here, we repeat this study with  $m_h=90$ and, for comparison, with $m_h=120$. The resulting estimation errors for the three $\hat{K}$ DNN sizes are given in Table \ref{table:rand_nn}.
In this comparison study, we fix the $\hat{h}$ and $\hat{C}$ DNNs' size at $m_h = 60$ and use $N_K = 80$, $N_h = 40$, and $N_C = 100$ measurements and  $N_f^h = 1000$ and $N_f^C = 1000$ residual points. We can see that the optimal-size $\hat{K}$ produces the smallest estimation errors not only for the $K$ field but also for the $h$ field in all the data-driven DNN, PINN-Darcy, and MPINN methods. For the $C$ field,  the smallest error is achieved with  $m_h=60$ in the  $\hat{K}$ DNN. This shows that a smaller estimation error in $K$ and $h$ does not always translate into a smaller error in $C$. 

\begin{table}[htb]
	\centering
	\caption{The relative errors of the estimated $K$, $h$, and $C$ for the $\hat{K}$ network size $m_h = 60, 90, 120$. The size of the $h$ and $C$ DNNS is fixed at $m_h = 60$. The DNN architecture for all three DNNS is [2-$m_h$-$m_h$-$m_h$-1].  $N_K = 80$, $N_h=40$, and $N_C = 100$ measurements and  $N_f^h = 1000$ and  $N_f^C = 1000$ residual points are used in this example. The log-conductivity field has the correlation length of $\lambda = 0.2$.}
	\begin{tabular}{c|ccc|ccc|ccc}
		\toprule
		\multicolumn{1}{c|}{} & \multicolumn{3}{|c}{$m_h=60$} & \multicolumn{3}{|c}{$m_h=90$} & \multicolumn{3}{|c}{$m_h=120$} \\
		\hline
		\multicolumn{1}{c|}{} 	& $\epsilon^K$    & $\epsilon^h$    & $\epsilon^C$   & $\epsilon^K$    & $\epsilon^h$    & $\epsilon^C$ & $\epsilon^K$    & $\epsilon^h$    & $\epsilon^C$ \\
		\hline
		\multicolumn{1}{c|}{DNN}       & $64.8 \%$  &                    &                & $54.2 \%$   &                     &                   & $60.5\%$  &                    &        \\
		\multicolumn{1}{c|}{PINN} 	    & $49.2 \%$  & $4.35 \%$   &                  & $48.5 \%$    & $3.95 \%$   &                   & $51.7 \%$    & $4.40 \%$   &         \\
		\multicolumn{1}{c|}{MPINN}    & $41.9 \%$    & $3.74 \%$   & $7.10 \%$ & $40.2 \%$    & $3.64 \%$   & $11.3 \%$   & $53.1 \%$    & $4.02 \%$   & $8.95 \%$\\
		\bottomrule
	\end{tabular}
	\label{table:rand_nn}
\end{table}

%
%%%%%%%%%%%%%%%%%%%%%%%%%%%%%%%%%%%%%%%%%%%%%%%%%%%%
%%%     Conclusion     %%%
%%%%%%%%%%%%%%%%%%%%%%%%%%%%%%%%%%%%%%%%%%%%%%%%%%%%
\section{Conclusion}\label{sec:conclusions}
In this study, we presented a physics-informed neural network approach for data and model assimilation for parameter and state estimation in multiphysics problems with application to subsurface transport problems. In this approach, all unknown parameter fields and states are modeled with DNNs, which are jointly trained by minimizing the loss function containing the multiphysics data (e.g., conductivity, hydraulic head, and concentration measurements ) and the associated physical models constraints.  As a result, the DNNs can be trained using indirect measurements, which is important when the data is sparse.

By representing the space-dependent parameter and state variables with DNNs, the  physics-informed approach offers a flexible and unified framework to deal with sparse and multiphysics data. In this study, we have compared three approaches: 1) the pure data-driven DNN approach, which only uses data to train DNNs; 2) the physics-informed DNN approach, called "PINN-Darcy," which utilizes the conductivity and hydraulic head measurements and the Darcy equation; and 3) the physics-informed DNN approach,  which combines the conductivity, head, and concentration measurements with the Darcy and advection--dispersion equations.  In this work, we referred to the third approach as MPINN. Our numerical results show that MPINN improves the accuracy of PINN-Darcy estimates of conductivity, and both physics-informed methods (PINN-Darcy and MPINN) significantly improve the accuracy of the data-driven DNN conductivity estimation and reduce the uncertainty in the DNN prediction, especially when the  direct measurements are limited.

We investigated the effect of the neural network size on the accuracy of parameter and state estimation as a function of the correlation length of the modeled $K$ field.
We demonstrated that  small and large networks might result in poor representability or overfitting of DNNs and identified an optimal DNN size that has a power-law dependence on the correlation length.   
 The physics constraints and added measurements reduce dependence of the DNN prediction on the DNN size given that the DNN is large (representative) enough. However, for a small number of measurements, we demonstrated that an optimal-size DNN outperforms the larger and smaller DNNs.

In subsurface applications, data is usually sparse and is often indirect. Therefore, the physics-informed DNNs provide an attractive alternative to standard data-driven machine learning methods.   Since the proposed method involves training DNNs by minimizing the loss function, the performance of training algorithms is crucial. In our study, we found that introducing nonlinear PDE constraints into the loss function increases the computational cost of training. Application of the physics-informed DNNs  to large-scale problems will require access to multi-GPU computers and scalable training algorithms. The selection of training algorithms and hyperparameters (width and depth of DNNs) should also be studied in more details.

%
%%%%%%%%%%%%%%%%%%%%%%%%%%%%%%%%%%%%%%%%%%%%%%%%%%%%
%%%     End of body of article     %%%
%%%%%%%%%%%%%%%%%%%%%%%%%%%%%%%%%%%%%%%%%%%%%%%%%%%%
\section*{Acknowledgements}
This research was partially supported by the U.S. Department of Energy (DOE) Advanced Scientific Computing (ASCR) program. PNNL is operated by Battelle for the DOE under Contract DE-AC05-76RL01830. 
The data and codes used in this paper are available at https://github.com/qzhe-mechanics/DataAssi-transport.

%\section*{References}
\bibliography{PINN_Ref}  %%% Remove comment to use the external .bib file (using bibtex).

%\begin{thebibliography}{1}
%\bibitem{kour2014real}
%George Kour and Raid Saabne.
%\newblock Real-time segmentation of on-line handwritten arabic script.
%\newblock In {\em Frontiers in Handwriting Recognition (ICFHR), 2014 14th
%  International Conference on}, pages 417--422. IEEE, 2014.
%\end{thebibliography}

\end{document}